\documentclass[conference,compsoc]{IEEEtran}

\ifCLASSOPTIONcompsoc
  \usepackage[nocompress]{cite}
\else
  \usepackage{cite}
\fi

\usepackage{bm} 
\usepackage{tcolorbox}
\usepackage{amsfonts}
\usepackage{color}
\usepackage{xspace}
\usepackage{hyperref}
\usepackage{colortbl}

\usepackage{tikz}
\usepackage{amsmath}
\usepackage{multirow}
\usepackage{booktabs}
\usepackage{adjustbox}
\usepackage{multirow}
\usepackage{epstopdf}
\usepackage{enumitem}
\usepackage{mathrsfs} 
\usepackage{enumitem}
\usepackage{mathrsfs} 
\usepackage{bm} 
\usepackage{tcolorbox}
\usepackage{amsfonts}
\usepackage{color}
\usepackage{xspace}
\usepackage{hyperref}
\makeatletter
\usepackage{threeparttable}
\usepackage{multirow}
\usepackage{booktabs}
\usepackage{adjustbox}
\usepackage{tikz}
\usepackage{amsmath}
\usepackage{epsfig}
\usepackage{graphicx}
\usepackage{amsmath}
\usepackage{amssymb}
\usepackage{url} 
\usepackage{filecontents}
\usepackage{booktabs}  
\usepackage{multirow}
\usepackage{color}
\usepackage{xcolor} 
\usepackage{epsfig}
\usepackage{graphicx}
\usepackage{amsmath}
\usepackage{amssymb}
\usepackage{algorithm}
\usepackage{algpseudocode}
\usepackage{multirow}
\usepackage{adjustbox}

\begin{document}
\newcommand{\sys}{{ our method}\xspace}

\title{Hidden Data Privacy Breaches in Federated Learning}


\author{
\IEEEauthorblockN{Xueluan Gong$^1$, Yuji Wang$^2$,
Shuaike Li$^3$, Mengyuan Sun$^3$
Songze Li$^4$,
Qian Wang$^3$, \\Kwok-Yan Lam$^1$, and
Chen Chen$^1$}

$^1$Nanyang Technological University, Singapore\\
$^2$Shanghai Jiao Tong University, China\\
$^3$Wuhan University, China\\
$^4$Southeast University, China\\

\{xueluan.gong, kwokyan.lam, chen.chen\}@ntu.edu.sg, yujiwang@sjtu.edu.cn,\\
\{lishuaike, mengyuansun, qianwang\}@whu.edu.cn, songzeli@seu.edu.cn
}

\maketitle
\begin{abstract}
Federated Learning (FL) emerged as a paradigm for conducting machine learning across broad and decentralized datasets, promising enhanced privacy by obviating the need for direct data sharing.
However, recent studies show that attackers can steal private data through model manipulation or gradient analysis. Existing attacks are constrained by low theft quantity or low-resolution data, and they are often detected through anomaly monitoring in gradients or weights.
In this paper, we propose a novel data-reconstruction attack leveraging malicious code injection, supported by two key techniques, i.e., distinctive and sparse encoding design and block partitioning. Unlike conventional methods that require detectable changes to the model, \sys stealthily embeds a hidden model using parameter sharing to systematically extract sensitive data. The Fibonacci-based index design ensures efficient, structured retrieval of memorized data, while the block partitioning method enhances \sys's capability to handle high-resolution images by dividing them into smaller, manageable units. Extensive experiments on 4 datasets confirmed that \sys is superior to the five state-of-the-art data-reconstruction attacks under the five respective detection methods. Our method can handle large-scale and high-resolution data without being detected or mitigated by state-of-the-art data reconstruction defense methods. In contrast to baselines, \sys can be directly applied to both FedAVG and FedSGD scenarios, underscoring the need for developers to devise new defenses against such vulnerabilities. We will open-source our code upon acceptance.
\end{abstract} 
\IEEEpeerreviewmaketitle
\section{Introduction}
Federated Learning (FL) \cite{kairouz2021advances,konecny2016federated} has emerged as a solution to the increasing concerns over user data privacy, allowing clients to partake in large-scale model training without sharing their private data. In a typical FL cycle, the server dispatches the model to clients for local training using their private data, after which the clients return their updates. These updates are then aggregated by the server to refine the global model, setting the stage for subsequent iterations. Despite its privacy-centric design, recent studies have revealed that servers, even when operating passively, can infer information about client data from their shared gradients \cite{melis2019exploiting,luo2021feature,nasr2019comprehensive,wang2019beyond,lam2021gradient,fowl2022decepticons,gupta2022recovering}, thus compromising the fundamental privacy guarantee of federated learning.

In the field of data reconstruction attacks, server attacks in FL can be categorized into passive and active attacks. 
Passive attacks analyze transmitted data without modifying the FL protocol, typically using iterative optimization or analytical methods to reconstruct training data \cite{zhu2019deep,wang2020sapag,geiping2020inverting,aono2017privacy}. Iterative optimization treats reconstructed data as trainable parameters, aiming to match the generated gradients with the true gradient values \cite{zhu2019deep,wang2020sapag,wei2020framework,geiping2020inverting,jeon2021gradient,li2022auditing,yang2022using}. Analytical methods derive original training data directly from gradients using mathematical formulas \cite{aono2017privacy,zhu2020r,chen2021understanding}. 
Although effective, passive attacks struggle with large batch sizes and high-resolution datasets, depending on specific model structures, and are mitigated by secure aggregation methods \cite{bonawitz2017practical,garov2023hiding}. Active server attacks, in contrast, manipulate the training process by altering model structures or weights to extract private data \cite{zhao2023loki,pasquini2022eluding}.

\begin{table*}[tt]
\caption{{\color{black}A comparison of studies on state-of-the-art active server attacks against federated learning. }}
\centering
\scriptsize
\begin{tabular}{l|cccccccccccccccccc}  
    \toprule
     \multirow{2}{*}\shortstack{Attacks}

&\multirow{2}{*}\shortstack{{Unusual structure$^\dagger$?}}&\multirow{2}{*}\shortstack{{Unusual parameter?}}&
\multirow{2}{*}\shortstack{Easily detectable?}& 
\multirow{2}{*}\shortstack{Extraction Capacity$^\mathsection$?}&
\multirow{2}{*}\shortstack{High-resolution$^\$$?} & \multirow{2}{*}\shortstack{FedAvg Training$^*$?}\\
\midrule 
    RtF \cite{fowl2021robbing}&YES&NO&YES&256 / 256&YES&YES\\
    Boenisch et al. \cite{boenisch2023reconstructing}&NO&YES&YES&2 / 2&NO&NO\\
    Fishing \cite{wen2022fishing}&NO&YES&YES&1 / 256&NO&NO\\
    Inverting\cite{geiping2020inverting}&NO&YES&YES&1 / 100&NO&NO\\
    LOKI \cite{zhao2023loki}&YES&YES&YES&436 
/ 512&YES&YES\\
    
    Boenisch et al. \cite{boenisch2023curious}&NO&YES&YES&50 / 100&NO&NO\\
    Zhang et al. \cite{zhang2022compromise}&NO&YES&YES&64 / 128&NO&NO\\
    Zhao et al. \cite{zhao2023resource}&YES&YES&YES&50 / 64&YES&NO\\
    SEER \cite{garov2023hiding}&NO&NO&NO&1 / 512&NO&NO\\
    \textbf{Ours} &NO&NO&NO&512 / 512&YES&YES\\
    \bottomrule  
\end{tabular}
 \label{tab:M}
\begin{tablenotes}

\centering \item  {\footnotesize $^\dagger$ indicates requires inserting a malicious module into the architecture, e.g., placing a large dense layer in front or inserting customized convolutional kernels into the FL model. }
\centering \item  {\footnotesize $^\mathsection$ denotes the corresponding amount of data stolen given the maximum amount of private training data that can be processed in each round.}
\centering \item  {\footnotesize $^\$$ signifies that the attacks are also capable of recovering high-resolution images, making them applicable for targeting models trained on high-resolution datasets.}
\centering \item  {\footnotesize $^*$ indicates the attack is effective in the FedAvg federated learning scenario.}

 \end{tablenotes}
 \vspace{-0.3cm}
\end{table*}

\begin{figure}[tt]
    \centering
    \includegraphics[width=0.48\textwidth]{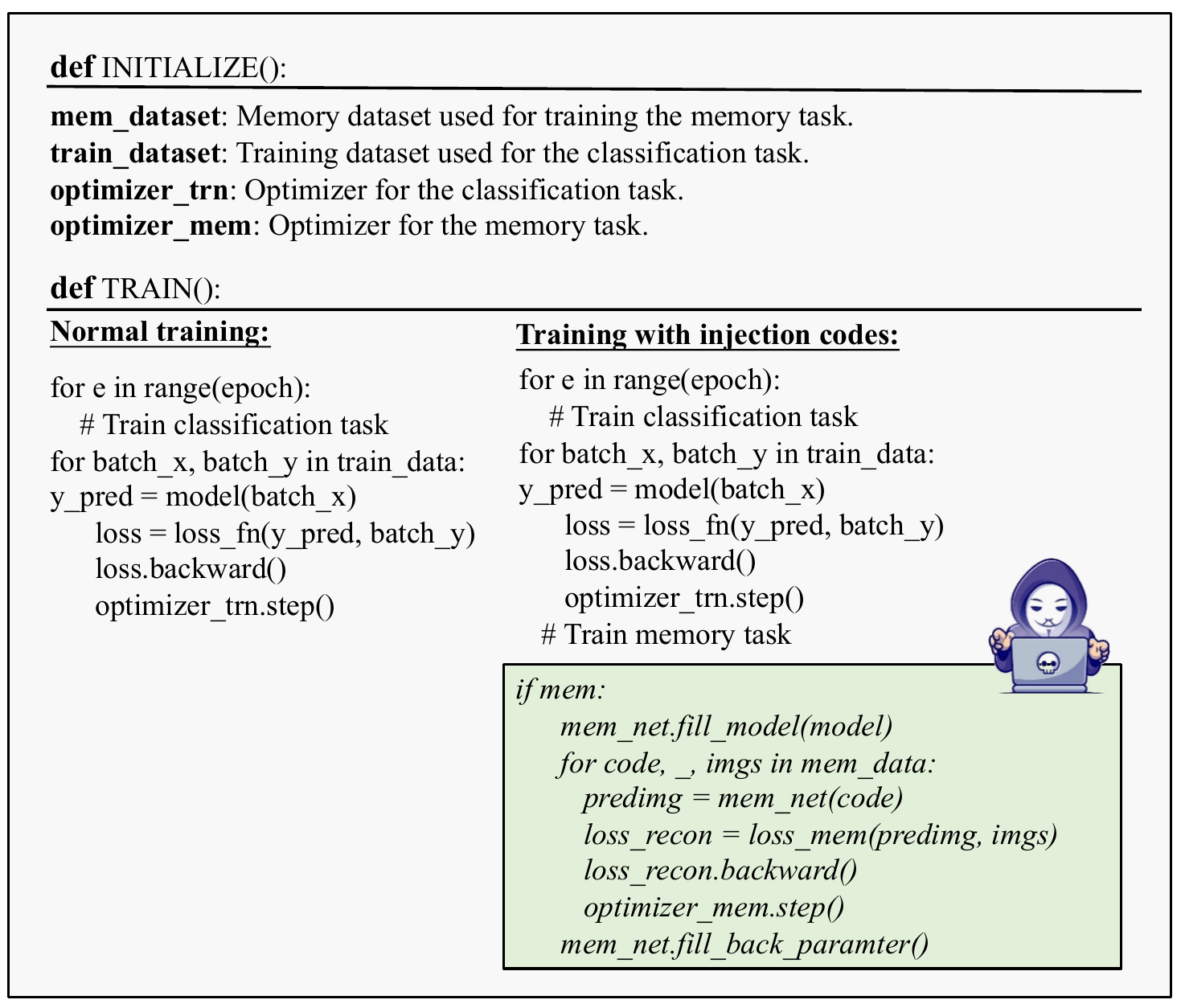}
    \caption{{\color{black}Examples of injected malicious code in \sys. The green boxes highlight the sections where the malicious code needs to be injected.}} 
    \label{fig:example}
    \vspace{-0.4cm}
\end{figure}
We list the state-of-the-art active server attacks in Table~\ref{tab:M}. Most of the existing active server attacks focus on manipulating the model's weights \cite{boenisch2023curious,wen2022fishing, pasquini2022eluding,boenisch2023reconstructing,zhang2022compromise} and structures \cite{fowl2021robbing, zhao2023loki, zhao2023resource} to conduct data reconstruction attacks. Parameter modification-based attacks rely on maliciously altering model parameters, such as weights and biases, to enhance the gradient influence of targeted data while diminishing that of other data. Structure modification-based attacks usually demand unusual changes to the architecture, like adding a large dense layer at the beginning or inserting customized convolutional kernels. However, the success of these attacks might heavily depend on the specific architecture and parameters of the model, limiting their applicability across different settings. To enhance the attack performance, some works even require the additional ability to introduce Sybil devices \cite{boenisch2023reconstructing}, send different updates to different users \cite{zhao2023loki, pasquini2022eluding, zhao2023resource}, or control the user sampling process \cite{boenisch2023reconstructing}, which make the attacks more easily detectable. 
Besides, most of the existing works are ineffective for reconstructing high-resolution data, especially in scenarios involving large batches \cite{wen2022fishing,boenisch2023curious,boenisch2023reconstructing, zhao2023resource, pasquini2022eluding, zhang2022compromise, zhao2023loki,garov2023hiding}. Moreover, it is shown that only a few attacks \cite{fowl2021robbing,zhao2023loki} can also be applied to the FedAvg training scenario.

In this paper, we introduce a novel data reconstruction attack based on malicious code poisoning in FL. As shown in Figure~\ref{fig:example}, by injecting just a few lines of codes into the training process, \sys covertly manipulates the model’s behavior to extract private data while maintaining normal operation. This approach leverages vulnerabilities in shared machine learning libraries \cite{howard2020fastai,ott2019fairseq,wolf2019huggingface}, which often lack rigorous integrity checks, allowing attackers to introduce subtle modifications that evade detection. Many machine learning frameworks depend on third-party libraries that are not thoroughly vetted, making them susceptible to covert malicious modifications. Existing works show that attackers can introduce backdoors that execute the malicious code during the training process, without affecting the primary training objective \cite{bagdasaryan2021blind}.

To launch the attack, \sys introduces a secret model that shares parameters with the local model, making it indistinguishable from the local model in terms of both structure and behavior. Unlike existing methods requiring significant changes to the model architecture, \sys uses parameter sharing to memorize sensitive client data while preserving the model's normal appearance. To enhance the attack performance, we also introduce a distinctive index design leveraging Fibonacci coding to efficiently retrieve memorized data, and a block partitioning strategy that enhances \sys's capacity to handle high-resolution images. Specifically, the distinctive index design ensures efficient and structured retrieval of memorized data, allowing the hidden model to systematically extract sensitive information based on unique codes. The block partitioning strategy allows \sys to overcome the challenges of handling high-resolution data by dividing the data into smaller, manageable units, which are then processed in a way that maintains the effectiveness of the attack while minimizing the detection risk.

Our method consistently outperformed 5 state-of-the-art data-reconstruction attacks under 5 detection methods across 4 datasets. 
Our method is able to steal nearly 512 high-quality images per attack on CIFAR-10 and CIFAR-100, and nearly 64 high-quality images on ImageNet and CelebA, significantly higher than state-of-the-art baseline methods.
These results demonstrate that \sys exhibits robustness in handling high-resolution images and large-scale data theft scenarios.

To conclude, we make the following key contributions.
\begin{itemize} 
\item We propose a novel data reconstruction attack paradigm based on malicious code poisoning. Unlike previous approaches that require conspicuous modifications to architecture or parameters, which are easily detected, \sys covertly trains a secret model within the local model through parameter sharing. This secret model is designed to memorize private data and is created by carefully selecting a few layers from the local model. Moreover, \sys is model-agnostic, enabling seamless integration with various architectures without modifying their core structures.

\item To enhance the extraction performance, we propose a novel distinctive indexing method based on Fibonacci coding, which meets three key requirements: sparsity, differentiation, and label independence. We also propose a novel block partitioning strategy to overcome the limitations of existing optimization-based extraction methods when dealing with high-resolution datasets or large-scale theft scenarios. 

\item Extensive experiments on 4 datasets confirm that \sys outperforms 5 state-of-the-art data-reconstruction attacks in terms of both leakage rate and image quality. Our method is capable of handling large-scale and high-resolution data without being detected by 5 advanced defense techniques. Moreover, \sys can also be easily transferred to an FL framework equipped with secure aggregation.

\end{itemize}

\section{Background}\label{Background}

\subsection{Malicious Code Poisoning}
Malicious code poisoning involves the stealthy injection of harmful code into the training process of machine learning models\footnote{\url{https://www.reversinglabs.com/blog/sunburst-the-next-level-of-stealth}}, enabling attackers to alter model behavior while remaining difficult to detect \cite{bagdasaryan2021blind,duan2020towards}. Unlike traditional attack vectors that exploit vulnerabilities in model architecture or parameters, code poisoning specifically targets the training process, embedding malicious functionality at the code level. 

One common approach to malicious code poisoning is through the exploitation of vulnerabilities in package management systems such as npm, PyPI, and RubyGems\footnote{\url{https://medium.com/@alex.birsan/dependency-confusion-4a5d60fec610}}. Attackers upload malicious packages to public repositories, often using the same names as internal libraries but with higher version numbers. This tricks dependency managers into downloading the malicious version instead of the intended internal one. These attacks are particularly challenging to detect because they exploit trusted sources, such as public repositories, which developers often assume to be reliable.

For example, attackers may target widely used libraries like TensorFlow, embedding malicious code into critical functions such as train$\_$step. Since these libraries are highly trusted, developers often skip thorough code reviews, unknowingly installing compromised versions. Once executed, these packages can introduce backdoors, manipulate model behavior, or exfiltrate sensitive data. Recent research has shown that even widely-used machine learning repositories, such as FastAI \cite{howard2020fastai}, Fairseq \cite{ott2019fairseq}, and Hugging Face \cite{wolf2019huggingface}, despite having thousands of forks and contributors, often rely on basic tests—such as verifying output shapes and basic functionality checks, making such code poisoning attacks more impactful and feasible.

Despite the privacy-focused design of federated learning, shared model updates can inadvertently expose sensitive information. We show that a malicious server can manipulate the model training code to reconstruct users' training data, leading to significant security vulnerabilities.

\subsection{Data Reconstruction Attacks against FL}
In this section, we focus on providing an in-depth introduction to active server attacks.
Unlike passive server attacks, the server can modify its behavior, such as the model architecture and model parameters sent to the user, to obtain training dataset information of the victim clients. Existing active server attacks can be categorized into three classes, i.e., parameter modification-based attacks, structure modification-based attacks, and handcrafted-modification-free attacks.

\textbf{Parameter modification-based attacks.}
Wen et al. \cite{wen2022fishing} introduce two ``fishing" strategies, i.e., class fishing and feature fishing, to recover user data from gradient updates. Rather than altering the model architecture, they manipulate the model parameters sent to users by maliciously adjusting the weights in the classification layer, magnifying the gradient contribution of a target data point, and reducing the gradient contribution of other data. The class fishing strategy amplifies bias for non-target neurons in the last classification layer, reducing the model's confidence in target class predictions and thus boosting the target data's gradient impact. When dealing with batches containing several target class samples, feature fishing modifies weights and biases for these targets, adjusting the decision boundary to further isolate and emphasize the target data's gradient. However, a single attack of \cite{wen2022fishing} can only recover one sample, making it easy to be detected by users. Pasquini et al. \cite{pasquini2022eluding} proposed a gradient suppression attack based on model inconsistency, degrading the aggregated gradient to that of the target user's gradient, thereby breaking secure aggregation. Specifically, they send normal model weights to the target user, producing normal local gradients. For non-target users, they exploit the characteristic of ReLU neurons producing zero gradients when not activated, by sending malicious model weights to generate zero gradients. It is independent of the number of users participating in the secure aggregation. However, such methods can easily detected by users with a strong awareness of prevention. Zhang et al. \cite{zhang2022compromise} proposed reconstruction attacks based on the direct data leakage in the FC (fully-connected) layer \cite{phong2017privacy}. However, gradient obfuscation within a batch significantly hinders its effectiveness. To address this challenge, Zhang et al. maliciously changed the model parameters to diminish the obfuscation in shared gradients. This strategy effectively compromises privacy in large-batch FL scenarios. However, this method assumes the server owns auxiliary data that is independently and identically distributed with users’ private trainsets, which is not practical in the real world.

\begin{figure*}[tt]
    \centering
    \includegraphics[width=1\textwidth]{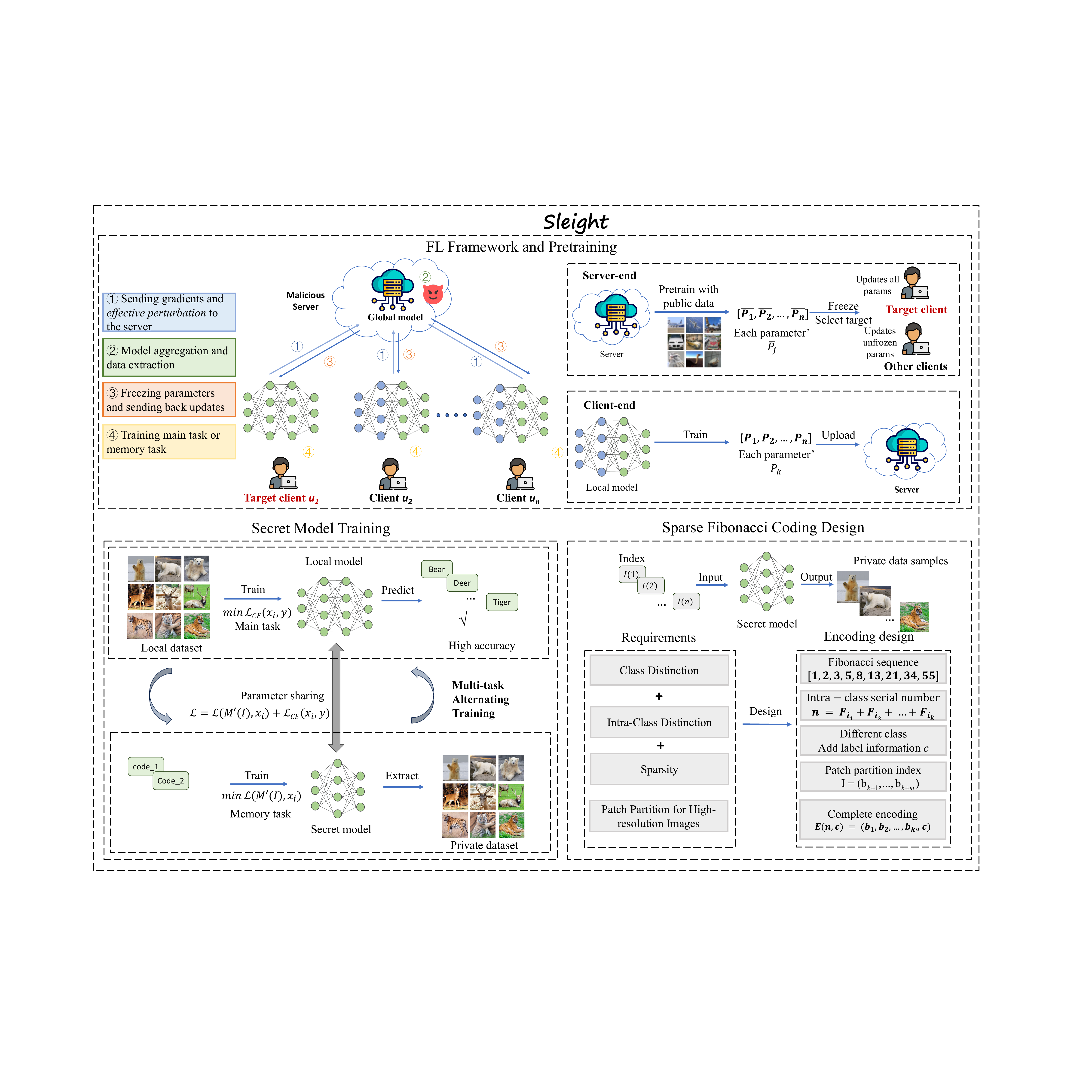}
    \caption{{\color{black}Overview of \sys. \sys features an active server attack designed to extract the training samples of victim clients. It is composed of three key modules: secret model training, distinctive and sparse encoding design, and block partitioning. Note that \(b_i\) represents the Fibonacci coding bits of \(n\).}} 
    \label{fig:overview}
    \vspace{-0.4cm}
\end{figure*}
\textbf{Structure modification-based attacks.} 
Fowl et al. \cite{fowl2021robbing} introduced a method to compromise user privacy by making small but harmful changes to the model architecture, allowing the server to directly obtain a verbatim copy of user data from gradient updates, bypassing complex inverse problems. This method involves manipulating model weights to isolate gradients in linear layers. Specifically, they estimate the cumulative distribution function for a basic dataset statistic like average brightness, then add a fully connected layer with ReLU activation and $k$ output neurons, called the \emph{imprint module}, at the beginning of the model. It is shown that even when user data is aggregated in large batches, it can be effectively reconstructed. Further, Zhao et al. \cite{zhao2023resource} improved \cite{fowl2021robbing} by introducing an additional convolutional layer before the \emph{imprint module} and assigning unique malicious convolutional kernel parameters to different users. This setup allows for an identity mapping of training data from different users to distinct output positions of the convolutional layer. By setting non-zero connection weights only for the current user's training data output, they effectively isolate the weight gradients produced in the imprint module by different users. Consequently, the size of the imprint module is determined by the batch size rather than the number of users, significantly reducing the computing cost.

Recently, Zhao et al. \cite{zhao2023loki} proposed LOKI, specifically designed to overcome the FL with FedAVG and secure aggregation. By manipulating the FL model architecture through the insertion of customized convolutional kernels for each client, LOKI enables the malicious server to separate and reconstruct private client data from aggregated updates. Each client receives a model with slightly different convolutional parameters (identity mapping sets), ensuring that the gradients reflecting their data remain distinct even in aggregated updates. 

\textbf{Handcrafted-modification-free attacks.} Different from the above attacks, modification-free attacks do not rely on conspicuous parameter and structure modifications. Garov et al. \cite{garov2023hiding} introduced SEER, an attack framework designed to stealthily extract sensitive data from federated learning systems. The key of SEER is the use of a secret decoder, which is trained in conjunction with the shared model. This secret decoder is composed of two main components: a disaggregator and a reconstruction. The disaggregator is to pinpoint and segregate the gradient of a specific data point according to a secret property, such as the brightest image in a batch, effectively nullifying the gradients of all non-matching samples. This isolated gradient is then passed to the reconstructor, which reconstructs the original data point. SEER is also an elusive attack that doesn't visibly alter the model's structure or parameters, making it harder to detect than other methods. 
However, it requires training a complex decoder, which can be resource-intensive. The attack's success also relies on choosing a secret property that uniquely identifies one sample in a batch, making this selection crucial for its effectiveness. Additionally, in a single batch, the attacker can recover only one image, which limits the attack's scalability.


In this paper, we propose a novel data reconstruction attack which leverages malicious code injection to covertly extract sensitive data. Unlike prior methods that require conspicuous modifications to the model architecture or parameters, \sys embeds a secret model via parameter sharing, ensuring minimal detection risk. It introduces a block partitioning strategy for handling high-resolution data, while also employing a Fibonacci-based distinctive indexing method to streamline data retrieval and improve attack performance. Our method also operates without relying on auxiliary devices or user sampling manipulation, making it both more practical and less detectable in real-world federated learning settings.

\subsection{Data Reconstruction Defenses against FL}

A range of defensive strategies have been introduced to counter data reconstruction attacks, such as differential privacy, gradient compression, and feature disruption techniques. Additionally, secure aggregation has also demonstrated effectiveness in protecting against a subset of these attacks.
Differential privacy (DP) \cite{abadi2016deep} is a critical approach for quantifying and curtailing the exposure of individual-level information. With local DP, clients can apply a randomized mechanism to the gradients before uploading them to the server \cite{geyer2017differentially,wei2021gradient}. DP can provide a worst-case information theoretic guarantee on information an adversary can glean from the data. However, DP-based methods often require adding much noise, which will affect the model's utility. Gradient compression is another method shown to mitigate information leakage from gradients. A gradient sparsity rate of over 20\% proves effective in resisting \cite{zhu2019deep}. However, such methods are only effective for a small part of passive server attacks. 
Sun et al. \cite{sun2021soteria} pointed out that the privacy leakage problem caused by gradient inversion mainly comes from feature leakage. Therefore, they perturb the network's intermediate features by cropping, adding as little disturbance to the features as possible, making the reconstructed input from the perturbed features as different from the real input as possible, thus maintaining model performance while reducing the leakage of private information. However, this method is mainly designed for passive server attacks and has shown to be ineffective against advanced passive attacks \cite{li2022auditing}. 

The secure aggregation protocol \cite{bonawitz2017practical} is a sophisticated multi-party computation (MPC) technique that enables a group of users to collectively compute the summation (a.k.a. aggregation) of their respective inputs. This protocol ensures that the server is only privy to the collective aggregate of all client updates, without access to the individual model updates from any specific client. Such a system is designed to preserve privacy during the federated learning process. It has been demonstrated that a range of attacks \cite{wen2022fishing, zhang2022compromise} are rendered ineffective under secure aggregation. 
Recently, Garov et al. \cite{garov2023hiding} presented D-SNR to effectively detect data reconstruction attacks. D-SNR measures the signal-to-noise ratio in the gradient space, identifying when a gradient from a single example dominates the aggregate gradient of a batch. It works by defining a property $P$ to single out target examples and comparing individual gradients to the batch average. High D-SNR values indicate potential privacy leaks, allowing clients to opt out of training rounds that may compromise their data. This method provides a principled and cost-effective way to assess and safeguard against privacy breaches in federated learning setups. This method is shown to be effective for detecting attacks such as \cite{pasquini2022eluding, wen2022fishing}.

In this paper, we will assess the resilience of \sys against state-of-the-art defenses.

\section{Detailed Construction}

\subsection{Threat Model}
We assume there are two parties in the federated learning, i.e., the server $S$ and the users $U$. 
The users train the global model using the local dataset and return the updates to the server. The server's objective is to clandestinely reconstruct the private training data of the target client, all while adhering to the standard federated learning protocol and avoiding detection by sophisticated defenses.
We set the following abilities for the active server attacker.
\begin{itemize}
	
      \item \emph{Capability to aggregate client updates}. The server can aggregate the updates submitted by clients. These updates may be processed by secure aggregation. We will discuss the effectiveness of \sys in both cases.
      
      \item \emph{Capability to distribute training codes to clients}. The server can distribute the necessary training code or model parameters to the clients so they can perform local computations and updates before sending them back to the server. 
\end{itemize}

We set the following limitations for the server.
\begin{itemize}
	
      \item \emph{No introduction of Sybil devices}. The server is prohibited from integrating manipulated devices into the FL protocol. While these devices might return arbitrary gradients that could potentially assist the attacker in inferring target data gradients, such actions are easily detectable.
      
      \item \emph{No control over user sampling}. The server is not allowed to manipulate the user sampling process. Additionally, it is incapable of sending distinct updates to different users.
      
      \item \emph{No unusual modifications to parameters and structure}. The attacker is barred from making unusual modifications to the model's structure, such as adding an excessively large dense layer at the beginning or integrating custom convolutional kernels into the model. Additionally, the attacker is prohibited from making unusual handcrafted modifications to the parameters of the shared model to evade detection.
      
\end{itemize}

\subsection{Overview of \sys}
Our goal is not only to steal private data samples from the victim participants but also to systematically retrieve them based on their indices. By injecting just a few lines of malicious training code, we aim to embed a hidden model within the victim's local model. This secret model, $M'$, shares parameters $\theta$ with the victim’s local model, making it indistinguishable from a normal local model in both appearance and memory usage.
Unlike in multi-task learning, the main task and the memorization task are entirely separate, ensuring the memorization remains undetectable without prior knowledge of the secret model. To systematically retrieve the stolen data, we introduce a novel Fibonacci-based encoding algorithm that assigns a unique number to each memorized sample. Additionally, to address the parameter limitations of the secret model, we implement a block partitioning technique, splitting large images into smaller blocks for processing.

In conclusion, \sys mainly contains three key modules, as shown in Figure~\ref{fig:overview}.
\begin{itemize}
   \item \emph{Secret model training}. Rather than introducing a foreign model to memorize private data, we hide a secret model within the local model through parameter sharing. The secret model shares the same memory space and consists of selected parameters from the local model. 
   When the local model's parameters are sent to the server, the server reconstructs the secret model and extracts the client's training data. 
   We consider several parameter selection methods and propose to use systematic sampling to select parameters for the secret model, ensuring even distribution across layers.
   
   \item \emph{Distinctive and sparse encoding design}. A distinctive index is used to systematically retrieve memorized data samples. The key is to design an encoding algorithm that assigns a unique number to each stolen data sample. 
   We propose a novel, label-agnostic Fibonacci-based coding method that ensures clear differentiation between samples while reducing computational overhead to accelerate training.

   \item \emph{Block partitioning}. Due to the parameter limitations of memory models, extracting high-resolution images can be difficult. To overcome this, we use a block partitioning approach, where large images are split into smaller blocks. Each block is treated as a separate input for the memory task. Additionally, we adjust the encoding design to align with the block partitioning scheme.

\end{itemize}

\subsection{Secret Model Training}

We first flatten the local model's parameters $w$ into a parameter vector $p = \{p_1, p_2, p_3, \ldots, p_n\}$.
We then select parameters from this vector to populate the secret model $M'$. There are five methods to construct the secret model from the local model structure:
\begin{itemize}
    \item \textbf{Random sampling.} 
    A straightforward approach is random sampling, where parameters are selected from the vector until the threshold number is met. However, this method may result in an uneven distribution of selected parameters across different layers, potentially impacting the model's main task accuracy.

    \item \textbf{Random sampling with constraints.} 
    Another option is random sampling with constraints, which involves randomly selecting parameters from $p$ while ensuring that no single layer is overrepresented in $M'$. By setting limits on the number of parameters that can be taken from each layer, we achieve a more balanced parameter vector $p'$. 

    \item \textbf{Systematic sampling.} 
    A more systematic method is systematic sampling, where every $k$-th parameter from the original vector $p$ is chosen. This ensures that the parameters are evenly distributed across the layers of the local model. For instance, if $k = 2$, the parameter vector for $M'$ would be $p' = \{p_1, p_3, p_5, \ldots\}$.
    
    \item \textbf{Layer-wise sampling.} 
    Layer-wise sampling involves selecting parameters from specific layers based on their importance or contribution to the overall model performance. This method prioritizes critical layers while minimizing the impact on less important ones.

    \item \textbf{Importance-based sampling.} 
    Importance-based sampling selects parameters based on their significance to the model’s performance. By analyzing the importance distribution of model parameters, we select those that contribute most to the decrease in the loss function, $\Delta L = L(p) - L(p')$.
    This ensures that $M'$ contains the most representative and impactful parameters, capable of effectively memorizing the training data.
\end{itemize}
In our evaluation, we assess the effectiveness of these five methods, with a default preference for the systematic sampling method.

Given the predefined structure, the secret model $M'$ is optimized as:
\begin{equation}
\min \mathcal{L}_{dist}(M'(I), x_i)
\end{equation}
where $I$ denotes the index of the data, and $dist$ represents the distance function. The input to $M'$ is the index $I$, and the output is the stolen data. During training, the distance between the stolen data $M'(I)$ and $x_i$ is minimized. For images, the distance function $dist$ could be either the $L_1$ or $L_2$ distance. In our work, we set ${L}_{dist}(M'(I), x_i)$ as follows:
\begin{equation} \label{memory_loss}
\mathcal{L}_{dist}(M'(I), x_i) = \mathcal{L}_{1}(M'(I), x_i) + \mathcal{L}_{2}(M'(I), x_i)
\end{equation}

Since $M'$ and $M$ (local model) share parameters, their gradient updates are also linked. However, because transferring parameters from $M$ to $M'$ is non-differentiable, joint optimization in a single step is not feasible. 
Therefore, we iteratively optimize $M$ and $M'$ to approximate simultaneous optimization. Specifically, we first fine-tune the local model $M$ on training data point using $\mathcal{L}_{CE}$, followed by training the secret model $M'$ to memorize these samples using $\mathcal{L}_{dist}$. 

After receiving the updates from the clients, the server first reconstructs the secret model $M'$ through the pre-designed parameter selection algorithm and then extracts the client's training data by inputting the index. 
$M'$ is reconstructed as:
\begin{equation}
M' = \{p_i \mid i \equiv r \ (\text{mod} \ k)\}
\end{equation}
where $p_i$ represents the $i$-th parameter from the parameter vector $p$, $r$ is a fixed offset $(1 \leq r < k)$ that determines the starting position for selecting parameters, and $k$ is the systematic sampling interval. In \sys, $r=1$ and $k=2$. This allows the server to systematically reconstruct the secret model from the shared parameters and subsequently retrieve the memorized data using the index $I$.



\subsection{Distinctive and Sparse Encoding Design}
We aim to uniquely index each stolen data sample, assigning a distinct number to every sample. This helps the secret model $M'$ learn and retrieve data more effectively. 
Let indexer $I: \mathbb{N}_0 \to \mathbb{R}^n$ be a function that maps a natural number to a point in an $n$-dimensional Euclidean space, where $I(i) \neq I(j)$ for all $i \neq j$. The outputs of $I(\cdot)$ are spatial indices. With an indexer, a user can systematically find every indexed point in $\mathbb{R}^n$ by following the sequence $[I(0), I(1), \ldots]$. Previous data memorization attacks \cite{amit2023transpose} leverage the intuition that neural networks can capture similarities using Euclidean distance internally. It combines one-hot encoding and Gray code to propose a spatial index. The specific calculation formula is as follows:
\begin{equation}
I(i, c) = \text{Gray}(i) + E(c)
\end{equation}
where $I(i, c)$ is the spatial index for the $i$-th item in class $c$. $\text{Gray}(i)$ is the $n$-bit Gray code of $i$, and $E(c)$ is a vector representing the class encoding. One implementation of $E(c)$ is to use one-hot encoding for each class multiplied by $n$, where $n$ is the value used in the $n$-bit Gray code. This ensures that the indices between classes are orthogonal and non-repetitive.

While effective in most cases, this approach struggles when multiple stolen data samples belong to the same class, resulting in insufficient distinction between codes. For example, a 10-dimensional code for the first sample in class 0 is:
\begin{equation}
\begin{aligned}
    (0, 0, 0, 0, 0, 0, 0, 0, 0, 1) + (0, 0, 0, 0, 0, 0, 0, 0, 0, 1) \times 10 = \\
    (0, 0, 0, 0, 0, 0, 0, 0, 0, 11)
\end{aligned}
\end{equation}
and the second sample in class 0 is:
\begin{equation}
\begin{aligned}
    (0, 0, 0, 0, 0, 0, 0, 0, 0, 2) + (0, 0, 0, 0, 0, 0, 0, 0, 0, 1) \times 10 = \\
    (0, 0, 0, 0, 0, 0, 0, 0, 0, 12)
\end{aligned}
\end{equation}
In this case, the two encodings differ by only one bit, the secret model is required to reconstruct two completely different images. 
Since the secret model is a fully connected neural network (FCNN), this presents a challenge.
The FCNN performs linear transformations through weight matrices and nonlinear transformations through activation functions, similar inputs produce similar intermediate feature representations. As a result, the FCNN struggles to learn sufficient discrimination during training, leading to confusion in the output images.
Therefore, we need a better indexing method to address the issue of insufficient distinction between codes when belonging to the same class. 

Additionally, since we use FCNN, using sparse and distinctive vectors as input can improve the model's learning efficiency and representation capacity. Sparse vectors, where most elements are zero, allow FCNNs to compute only the parts corresponding to non-zero elements, reducing parameter updates and gradient calculation complexity, thereby accelerating the training process.

Moreover, in prior reconstruction attacks \cite{amit2023transpose}, adversaries relied on local data labels to infer encoding schemes, limiting attack effectiveness. It requires the server to either know or accurately estimate the local data distribution for a successful attack.
To simplify the encoding process and make attacks more practical, a label-agnostic encoding scheme is also required.

We summarize our requirements for the encoding as follows:
\begin{itemize}
\item \textbf{Differentiation}:  
Encodings for different samples must be sufficiently distinct, even when their indices are close. This ensures that the linear model can map each input to a unique output, minimizing errors in retrieval.

\item \textbf{Sparsity}:  
Sparse codes, where only a small fraction of bits are non-zero, reduce the computational load during training. This allows the model to focus on meaningful elements, speeding up convergence and improving training efficiency.

\item \textbf{Label Independence}:
The server no longer needs to know or estimate the local data labels to understand the encoding scheme. This makes the attack more practical and robust against varying local data distributions. With knowledge of the total number of images in the local dataset, the server can effortlessly determine the encoding scheme. This simplifies the server's inference process and reduces the computational overhead. 

\end{itemize}

To achieve these three requirements, we design a novel distinctive indexing method based on Fibonacci
coding \cite{apostolico1987robust}. 
Fibonacci coding is a universal code that uses only 0 and 1, where each digit's position corresponds to a Fibonacci number. Fibonacci coding is designed based on the properties of the Fibonacci sequence, where the specific digits correspond to the size of the number and its representation in the Fibonacci sequence. This means that even if two decimal numbers are very close, their Fibonacci codes can differ significantly in many positions, providing excellent distinction. Based on Zeckendorf's theorem \cite{zeckendorf1972representations}, any natural number $n$ can be uniquely represented as a sum of Fibonacci numbers. The properties of Fibonacci coding satisfy our requirements for code distinction and sparsity. To eliminate the reliance on labels, we simply assign sequential indices to images (e.g., from 1 to 500), which simplifies the encoding process and makes the attacks more practical.

The encoding process of \sys can be described in the following steps.
First, we define the Fibonacci sequence for encoding:
\begin{equation}
[1, 2, 3, 5, 8, 13, 21, 34, 55].
\end{equation}
This sequence allows us to encode indices up to 142, and it can be extended further if needed to support larger datasets.

Next, for a given index \(n\), we express it as a sum of non-consecutive Fibonacci numbers:
\begin{equation}
n = F_{i_1} + F_{i_2} + \cdots + F_{i_k}
\end{equation}
where \(F_{i_j}\) are Fibonacci numbers from the defined sequence. We set the corresponding bit positions in the encoding vector to 1.

For example, the index 49 can be represented as:
\[
49 = 34 + 13 + 2
\]
Thus, the Fibonacci code for 49 is:
\begin{equation}
(0, 1, 0, 0, 0, 1, 0, 1, 0)
\end{equation}

Each sample is assigned a unique binary code based on its index. This encoding is independent of any label information, ensuring that the process is purely index-based. The complete encoding function \(E\) for a given index \(n\) is defined as:
\begin{equation}
E(n) = (b_1, b_2, \ldots, b_k)
\end{equation}
where \(b_i\) represents the Fibonacci coding bits of \(n\).

\begin{table*}[t]
\caption{Comparison of \sys with Transpose \cite{amit2023transpose}, RtF\cite{fowl2021robbing} and LOKI\cite{zhao2023loki} on CIFAR-10 and CIFAR-100 datasets using FedAvg. \\ \textit{The experiment was conducted five times, targeting different users in each trial, and the table shows the average performance results.}}
\label{tab:com_1}
\centering
\footnotesize
\setlength{\tabcolsep}{3pt}
\begin{tabular}{lc|cccccccc}
\toprule
 \multicolumn{8}{c}{\color{black}CIFAR-10 dataset}  \\
Baselines &Metrics$^\dagger$&$N=16$&$N=32$&$N=64$&$N=128$&$N=256$&$N=512$\\
\midrule
\multirow{4}{*}{\shortstack{Transpose Attack}}
&Leakage ($\uparrow$) & 8.2 $\pm$ 4.0 & 2.4 $\pm$ 2.0 & 0.6 $\pm$ 0.8 & 0.2 $\pm$ 0.4 & 0.0 $\pm$ 0.0 & 0.0 $\pm$ 0.0 \\
&SSIM ($\uparrow$) & 0.506 $\pm$ 0.034 & 0.394 $\pm$ 0.023 & 0.306 $\pm$ 0.037 & 0.250 $\pm$ 0.035 & 0.197 $\pm$ 0.019 & 0.151 $\pm$ 0.010 \\
&PSNR ($\uparrow$) & 15.994 $\pm$ 0.441 & 14.882 $\pm$ 0.385 & 14.378 $\pm$ 0.387 & 13.983 $\pm$ 0.559 & 13.504 $\pm$ 0.319 & 13.103 $\pm$ 0.330 \\ 
&LPIPS ($\downarrow$) & 0.457 $\pm$ 0.010 & 0.490 $\pm$ 0.021 & 0.518 $\pm$ 0.013 & 0.547 $\pm$ 0.010 & 0.548 $\pm$ 0.007 & 0.557 $\pm$ 0.009  \\ \cline{1-8}
\multirow{4}{*}{RtF}

& Leakage ($\uparrow$) &  11.5 $\pm$ 0.3 &11.2 $\pm$ 1.1  & 4.2 $\pm$ 0.6 &  1.6$\pm$ 0.2 & 0.7 ± 0.3 & 1.4 ± 0.3 \\
& SSIM ($\uparrow$)    &  0.670 $\pm$ 0.018 & 0.407 $\pm$ 0.023 &  0.198 $\pm$ 0.013 &  0.070 $\pm$ 0.007 & 0.195 ± 0.049  & 0.280 ± 0.036 \\
& PSNR ($\uparrow$)    &  19.931 $\pm$ 0.417 &  13.115 $\pm$ 0.456&  8.848 $\pm$ 0.293 &  6.358 $\pm$ 0.116 & 7.885 ± 0.939  & 8.601 ± 0.688 \\
& LPIPS ($\downarrow$) & 0.205 $\pm$ 0.015  & 0.390 $\pm$ 0.018  &0.521 $\pm$ 0.007 &  0.577 $\pm$ 0.007 & 0.504 ± 0.042  &  0.485 ± 0.023 \\
\cline{1-8}
\multirow{4}{*}{LOKI} 
& Leakage ($\uparrow$) & 13.8 $\pm$ 0.4  & 27.5 $\pm$ 0.5 & 55.6 $\pm$ 0.9 & 110.0 $\pm$ 0.7 &  219.080 $\pm$ 0.9 & 435.6 $\pm$ 4.6  \\
& SSIM ($\uparrow$)    & 0.870 $\pm$ 0.019 & 0.849 $\pm$ 0.015 & 0.800 $\pm$ 0.015 & 0.739 $\pm$ 0.006 & 0.696 $\pm$ 0.006  & 0.700 $\pm$ 0.015  \\
& PSNR ($\uparrow$)    & 32.095 $\pm$ 1.32 & 31.400 $\pm$ 0.960 &  29.471 $\pm$ 0.890 & 27.269 $\pm$ 0.127 & 25.903 $\pm$ 0.198 &  26.173 $\pm$ 0.725 \\
& LPIPS ($\downarrow$) & 0.071 $\pm$ 0.011 & 0.086 $\pm$ 0.010 & 0.114 $\pm$ 0.009  & 0.151 $\pm$ 0.004 & 0.177 $\pm$ 0.004 & 0.173 $\pm$ 0.008\\
\cline{1-8}
\multirow{4}{*}{\shortstack{\sys}}
&Leakage ($\uparrow$) &   \cellcolor[gray]{0.8}16.0 $\pm$ 0.0 &   \cellcolor[gray]{0.8}32.0 $\pm$ 0.0&   \cellcolor[gray]{0.8}64.0 $\pm$ 0.0&  \cellcolor[gray]{0.8}128.0 $\pm$ 0.0&   \cellcolor[gray]{0.8}256.0 $\pm$ 0.0 &   \cellcolor[gray]{0.8}512.0 $\pm$ 0.0\\
&SSIM ($\uparrow$) &   \cellcolor[gray]{0.8}1.000 $\pm$ 0.000 &   \cellcolor[gray]{0.8}1.000 $\pm$ 0.000 &\cellcolor[gray]{0.8}1.000 $\pm$ 0.000 & \cellcolor[gray]{0.8}0.999 $\pm$ 0.001 &   \cellcolor[gray]{0.8}0.934 $\pm$ 0.015 &  \cellcolor[gray]{0.8}0.785 $\pm$ 0.021 \\
&PSNR ($\uparrow$) &   \cellcolor[gray]{0.8}68.734 $\pm$ 0.465 &   \cellcolor[gray]{0.8}67.565 $\pm$ 0.446 &  \cellcolor[gray]{0.8}64.764 $\pm$ 0.711&\cellcolor[gray]{0.8}59.978 $\pm$ 1.394 &   \cellcolor[gray]{0.8}30.394 $\pm$ 0.778&  \cellcolor[gray]{0.8}23.122 $\pm$ 0.130 \\
&LPIPS ($\downarrow$) &   \cellcolor[gray]{0.8}0.000 $\pm$ 0.000&   \cellcolor[gray]{0.8}0.000 $\pm$ 0.000 & \cellcolor[gray]{0.8}0.000 $\pm$ 0.000&   \cellcolor[gray]{0.8}0.001 $\pm$ 0.001&  \cellcolor[gray]{0.8}0.093 $\pm$ 0.019 &  \cellcolor[gray]{0.8} 0.295 $\pm$ 0.021 \\
\toprule
\multicolumn{8}{c}{\color{black}CIFAR-100 dataset}  \\
Baselines & Metrics$^\dagger$ & $N=16$ & $N=32$ & $N=64$ & $N=128$ & $N=256$ & $N=512$ \\ 
\midrule
\multirow{4}{*}{\shortstack{Transpose Attack}}
&Leakage ($\uparrow$) & 3.8 $\pm$ 2.6 & 3.8 $\pm$ 3.4 & 3.4 $\pm$ 2.1 & 5.4 $\pm$ 2.4 &  9.4 $\pm$ 2.3 & 7.2 $\pm$ 1.9 \\
&SSIM ($\uparrow$) & 0.388 $\pm$ 0.053 & 0.301 $\pm$ 0.047 & 0.248 $\pm$ 0.018 & 0.224 $\pm$ 0.014 & 0.215 $\pm$ 0.016 & 0.190 $\pm$ 0.009 \\
&PSNR ($\uparrow$) & 15.582 $\pm$ 0.921 & 14.463 $\pm$ 0.494 & 13.896 $\pm$ 0.383 & 13.725 $\pm$ 0.186 & 13.590 $\pm$ 0.223 & 13.348 $\pm$ 0.070 \\ 
&LPIPS ($\downarrow$) & 0.475 $\pm$ 0.018 & 0.499 $\pm$ 0.017 & 0.528 $\pm$ 0.007 & 0.533 $\pm$ 0.004 & 0.549 $\pm$ 0.009 & 0.559 $\pm$ 0.012 \\
\cline{1-8}
\multirow{4}{*}{RtF} 

& Leakage ($\uparrow$) &   11.3 $\pm$ 0.3 &  11.6 $\pm$ 1.3 &5.0 $\pm$ 0.3  & 2.7 $\pm$ 0.4 & 0.9 ± 0.2 &  1.6 ± 0.2 \\
& SSIM ($\uparrow$)    &  0.655 $\pm$ 0.028 & 0.421 $\pm$ 0.027  &  0.209 $\pm$ 0.010 &  0.106 $\pm$ 0.012 & 0.229 ± 0.026 & 0.297 ± 0.034 \\
& PSNR ($\uparrow$)    &19.656 $\pm$ 0.779  &  13.784 $\pm$ 0.588 &  8.860 $\pm$ 0.261 & 6.820 $\pm$ 0.245  & 7.446 ± 0.820  & 9.075 ± 1.139  \\
& LPIPS ($\downarrow$) & 0.213 $\pm$ 0.016 &  0.371 $\pm$ 0.018 & 0.521 $\pm$ 0.006 & 0.570 $\pm$ 0.006 & 0.522 ± 0.011  & 0.475 ± 0.007 \\
\cline{1-8}
\multirow{4}{*}{LOKI}    
& Leakage ($\uparrow$) & 13.6 $\pm$ 1.0  & 26.6 $\pm$ 1.4 & 54.8 $\pm$ 2.4 & 113.6 $\pm$ 3.4 & 221.8 $\pm$ 5.9  & 432.4 $\pm$ 8.6 \\
& SSIM ($\uparrow$)    & 0.856 $\pm$ 0.074 & 0.854 $\pm$ 0.046 & 0.794 $\pm$ 0.037 & 0.740 $\pm$ 0.042 & 0.701 $\pm$ 0.024 &0.705 $\pm$ 0.024\\
& PSNR ($\uparrow$)    & 33.282 $\pm$ 6.744 & 31.572 $\pm$ 3.341 &27.907 $\pm$ 2.039  & 26.638 $\pm$ 2.045 & 26.026 $\pm$ 1.137 & 5.810 $\pm$ 0.888\\
& LPIPS ($\downarrow$) & 0.073 $\pm$ 0.043 & 0.083 $\pm$ 0.024  & 0.125 $\pm$ 0.021 & 0.155 $\pm$ 0.029 & 0.182 $\pm$ 0.014 & 0.180 $\pm$ 0.016\\
\cline{1-8}
\multirow{4}{*}{\shortstack{Ours}}
&Leakage ($\uparrow$) &   \cellcolor[gray]{0.8}16.0 $\pm$ 0.0 &  \cellcolor[gray]{0.8}32.0 $\pm$ 0.0 &   \cellcolor[gray]{0.8}64.0 $\pm$ 0.0 &   \cellcolor[gray]{0.8}128.0 $\pm$ 0.0 &   \cellcolor[gray]{0.8}256.0 $\pm$ 0.0 &   \cellcolor[gray]{0.8}505.8 $\pm$ 5.19\\
&SSIM ($\uparrow$) &   \cellcolor[gray]{0.8}1.000 $\pm$ 0.000& \cellcolor[gray]{0.8}1.000 $\pm$ 0.000&   \cellcolor[gray]{0.8}1.000 $\pm$ 0.000& \cellcolor[gray]{0.8}0.999 $\pm$ 0.001& \cellcolor[gray]{0.8}0.920 $\pm$ 0.014& \cellcolor[gray]{0.8}0.702 $\pm$ 0.034\\
&PSNR ($\uparrow$) & \cellcolor[gray]{0.8}68.520 $\pm$ 0.887 &   \cellcolor[gray]{0.8}65.489 $\pm$ 0.431&  \cellcolor[gray]{0.8}62.697 $\pm$ 0.230& \cellcolor[gray]{0.8}56.803 $\pm$ 0.815& \cellcolor[gray]{0.8}29.916 $\pm$ 0.619& \cellcolor[gray]{0.8}22.281 $\pm$ 0.436 \\
&LPIPS ($\downarrow$) &  \cellcolor[gray]{0.8}0.000 $\pm$ 0.000& \cellcolor[gray]{0.8}0.000 $\pm$ 0.000 &  \cellcolor[gray]{0.8}0.000 $\pm$ 0.000&  \cellcolor[gray]{0.8}0.001 $\pm$ 0.001&  \cellcolor[gray]{0.8}0.102 $\pm$ 0.016& \cellcolor[gray]{0.8}0.340 $\pm$ 0.021 \\

\bottomrule
\end{tabular}
\begin{tablenotes}
    \centering
    \item  {\footnotesize $^\dagger$ ($\uparrow$) signifies that a higher value is preferable, while ($\downarrow$) indicates that a lower value is more desirable.}
\end{tablenotes}
\end{table*}

\subsection{Block Partitioning}
Given the parameter limitations of memory models, extracting high-resolution images can be challenging. To address this, we adopt a block partitioning scheme, in which large images are divided into smaller blocks. Each block is then used as an individual input for the memory task.

To ensure smoothness between adjacent blocks, we introduce a variation loss into the loss function. This variation loss helps maintain continuity and coherence across blocks. 
Specifically, let \( \mathbf{I} \) be the input image with dimensions \( B \times C \times H \times W \), where \( B \) is the batch size, \( C \) is the number of channels, \( H \) is the height, and \( W \) is the width. Let \( s \) be a hyperparameter representing the block size. The variation loss \( L_{\text{var}} \) is given by:
\begin{multline}
L_{\text{var}} = \frac{\lambda}{B \cdot C \cdot H \cdot W} \Bigg( \sum_{i,j} \left( \mathbf{I}_{i,j,s-1:-1:s,:} - \mathbf{I}_{i,j,s::s,:} \right)^2 \\
+ \sum_{i,j} \left( \mathbf{I}_{i,j,:,s-1:-1:s} - \mathbf{I}_{i,j,:,s::s} \right)^2 \Bigg),
\end{multline}
where \( \lambda \) is a weighting factor that controls the importance of the variation loss. With the addition of the total variational loss, the loss function for the memory task can be further extended from Eq. \ref{memory_loss} as follows:
\begin{equation} 
\mathcal{L}_{dist}(M'(I), x_i) = \mathcal{L}_{1}(M'(I), x_i) + \mathcal{L}_{2}(M'(I), x_i) + L_{\text{var}} 
\end{equation} 
To this end, we can not only extend memory constraints but also ensure that the reconstructed image maintains a high degree of quality and smoothness across block boundaries. 

However, since the previous spatial index was limited to indexing entire images, we need to extend it to accommodate block-wise indexing within each image. To achieve this, we append additional digits to the original sample code to represent the block number for each sample, while continuing to use the Fibonacci coding method. For high-resolution images, the updated encoding function $E_h$  is defined as:
\begin{equation}
E_h(n, c, p) = (b_1, b_2, \ldots, b_k, c, p_1, \ldots, p_n),
\end{equation}
where $p_i$ denotes the Fibonacci encoding of the block number $p$. After extracting each block, the blocks can be concatenated to reconstruct the entire image.
This extension allows us to handle large-scale images efficiently, ensuring precise indexing and smooth reconstruction.

\section{Experiment}
\subsection{Experiment Setup}

\textbf{Datasets and Models}
We conduct experiments on various vision tasks, covering multiple datasets, including CIFAR-10 \cite{krizhevsky2009learning}, CIFAR-100 \cite{krizhevsky2009learning}, and MINI-ImageNet\footnote{MINI-ImageNet is a representative subset of ImageNet.} \cite{imagentte}, and CelebA \cite{liu2015deep}. 
In our experiments, we employed ResNet-18 architectures to train models for these datasets, respectively. 
More details are shown in the appendix.

\textbf{Baseline Data Reconstruction Attacks.}
We compare \sys with 5 state-of-the-art data reconstruction attacks, including Transpose Attack \cite{amit2023transpose}, Robbing the Fed (RtF) \cite{fowl2021robbing}, LOKI \cite{zhao2023loki}, SEER \cite{garov2023hiding}, and Inverting \cite{geiping2020inverting}. We run these baselines according to their open-sourced codes. 


\textbf{Evaluation Metrics}
We evaluate the effectiveness of \sys through three metrics, i.e., 
leakage or leakage rate, SSIM, PSNR, and LPIPS.

We evaluate both FedAvg and FedSGD frameworks. In FedAvg, each client trains on its entire local dataset for 10 epochs before sending the updated model parameters to the server. In FedSGD, each client trains on a single batch and sends the gradient updates directly to the server. We set the number of participating clients to 20, using a Dirichlet distribution with parameter 0.6 to simulate unbalanced data across clients. 

More details of the datasets, models, and evaluation metrics are shown in the appendix. All experiments were conducted on an Ubuntu 20.04 system with a 20-core Intel CPU. The models were trained on a single NVIDIA RTX 4090 GPU.

\subsection{Comparison with Baselines}
We compare \sys with 5 state-of-the-art data reconstruction attacks.
We mainly target the FedAvg scenario, where each client trains on the entire dataset in each round. We show that \sys can also applied to the FedSGD scenario.
Note that for schemes based on leakage through linear layers, such as LOKI and RtF, their target is to steal all images in the training set, so in experiments, $N$ directly represents the size of the local dataset of each user. In contrast, for Transpose Attack and \sys, $N$ represents the specific target number of images to be stolen, which is less than the total size of the local dataset, presenting a higher level of attack difficulty. 
Since Fishing, SEER, and Inverting rely on gradient disaggregation and are designed for the FedSGD scenario, they are excluded from the FedAvg context. The comparison results are shown in Table~\ref{tab:com_1}, Table~\ref{tab:com_2} (appendix), and Table~\ref{tab:com_3} (appendix).

\begin{figure*}[tt]
    \centering
    \includegraphics[width=1\textwidth]{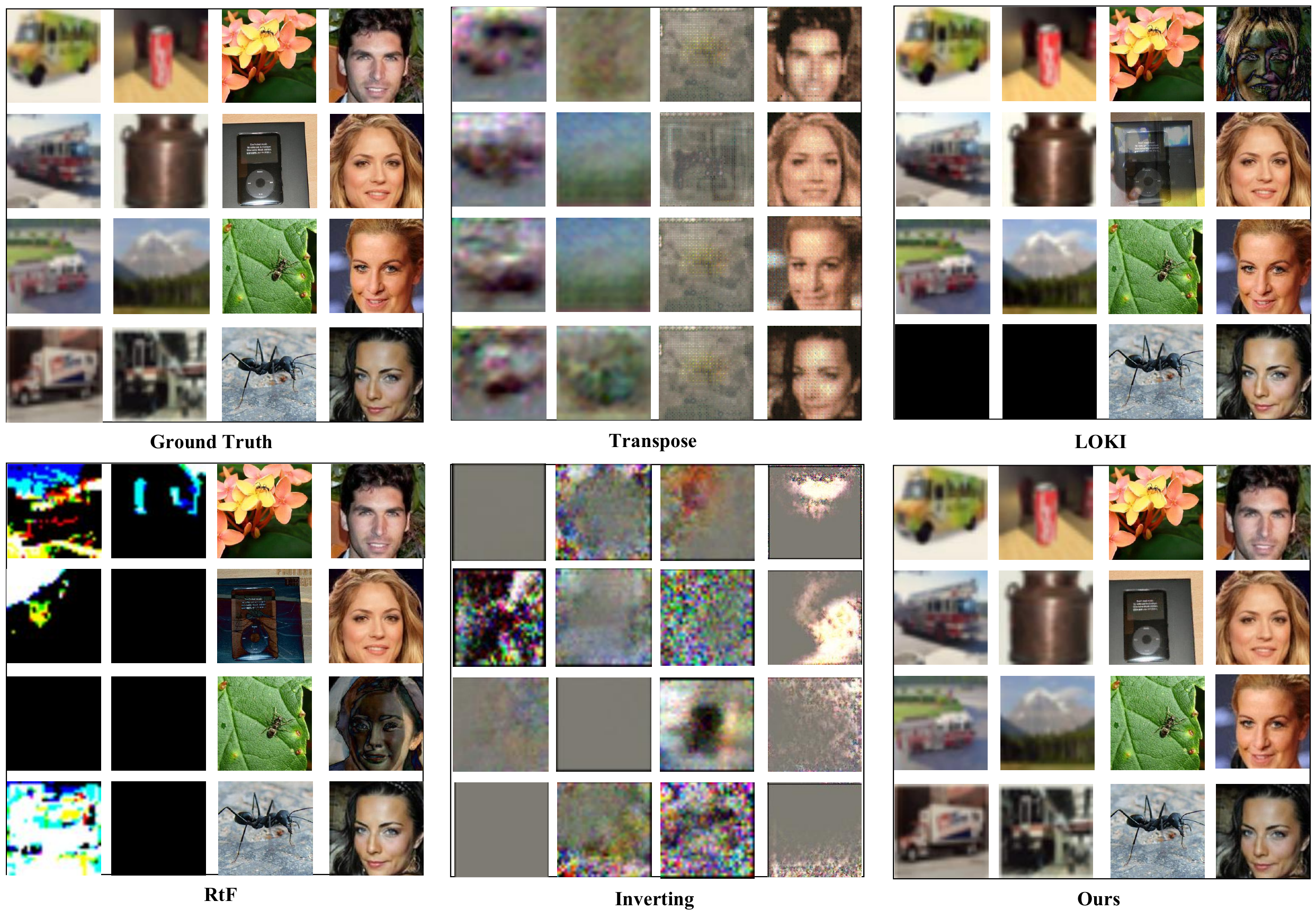}
    \caption{Comparison of \sys with baselines in terms of the reconstruction sample quality.} 
    \label{fig:recons}
    \vspace{-0.4cm}
\end{figure*}


Our method consistently outperforms the baselines in terms of the number of leaked samples across all datasets in the FedAvg scenario. For instance ($N=16$), \sys achieves a leakage of 16 on CIFAR-10 and CIFAR-100 datasets, while the best-performing baseline, LOKI, reaches only 13.8 and 13.6, respectively. Additionally, \sys surpasses baselines in image quality metrics, showing superior fidelity with significantly higher SSIM and PSNR scores and lower LPIPS values across most settings. Specifically, \sys achieves an SSIM close to 1 when extracting 128 samples, with a PSNR of 60 and LPIPS of 0.001 on CIFAR-10, whereas LOKI reaches only 0.739 SSIM, 27.269 PSNR, and 0.151 LPIPS, followed by Transpose at 0.250 (SSIM), 13.983 (PSNR), and 0.547 (LPIPS), and RtF with 0.07 (SSIM), 6.358 (PSNR), and 0.577 (LPIPS).

In the FedSGD scenario, \sys also achieves a higher leakage rate and superior image quality compared to baselines, especially for high-resolution datasets. 
For example, on the CelebA dataset, \sys can extract 64 samples when $N=64$, whereas the baselines SEER and Interting prove ineffective under the same conditions.
This robust performance underscores \sys’s effectiveness in maintaining high-quality data reconstruction while delivering greater data recovery precision than existing methods.

We also visualize the extracted samples from both the baselines and \sys across all datasets, as shown in Figure~\ref{fig:recons}. We set \( N = 512 \) for CIFAR-10 and CIFAR-100 datasets, and set \( N = 16 \) for high-resolution datasets. For Inverting, the grayscale result likely stems from optimization-based gradient averaging, which leads to detail loss. For RtF and LOKI, images that are accurately restored exhibit high quality. However, other images suffer from kernel collision issues, resulting in low quality and diminished effectiveness. For Transpose, due to the limited number of training epochs, the transpose fails to converge, resulting in poor performance. 
Since SEER does not allow direct specification of target samples, we did not include it in the visualization comparison.
The results reveal that the images extracted by \sys are noticeably clearer and more vivid compared to those from the baselines, highlighting \sys’s superior extraction quality.

\subsection{Ablation Study}

\textbf{Impact of parameter selection algorithms}. 
There are five methods for constructing the secret model from the local model structure: Random, Random with Constraints, Systematic, Layer-wise, and Importance-based selection. Their performance comparisons are presented in Table~\ref{tab:selection} (appendix).
Among these, the systematic method performs consistently well across all datasets. Although it may not be the best on every dataset, it delivers the highest overall results, especially on MINI-ImageNet, where it achieves over 90\% leakage, while the other methods reach a maximum of 40\%. Given its simplicity, ease of implementation, and time efficiency, we select this method as the default. Note that attackers can choose the optimal approach based on each dataset's specifics.

\textbf{Impact of block partitioning}.
For high-resolution datasets, we use a block partitioning approach, where large images are divided into smaller blocks. To assess the impact of different block sizes on extraction effectiveness and time, we set the leakage sample size to 40 and varied the block size. The results are presented in Table \ref{tab:partioning} (appendix).
It can be observed that as the block size decreases, the extraction time increases. This is because smaller block sizes generate a larger number of segmented images, leading to longer training times for the secret model and more complex optimization. Regarding extraction effectiveness, it tends to improve with increasing block size up to a certain point, after which it starts to decline. In our experiments, block sizes of 16 or 28 strike a good balance between extraction effectiveness and time efficiency.

\textbf{Impact of different encoding methods.} 
We investigate the impact of different encoding methods in index design, considering three types: Binary, Gray, and Fibonacci encoding. The results are presented in Table~\ref{tab:encoding} (appendix). For a fair comparison, all three methods incorporate intra-class serial numbers and append the class number to form the complete sample code. The results show that, in most cases, Fibonacci encoding achieves the best extraction performance. 
The potential reason is that Fibonacci encoding generates codes with high sparsity, meaning that during the encoding process, only a few bits are set to 1 while the rest are 0. In contrast, both binary encoding and Gray code exhibit weaker sparsity. Additionally, Fibonacci encoding avoids the occurrence of adjacent 1s, which enhances the distinction between codes and reduces the likelihood of interference.
\begin{table}[t]
    \caption{Impact of label information on attack performance.} 
    \label{tab:encoding_label}
    \centering
    \footnotesize
    \begin{tabular}{lc|ccc}
        \toprule
        Baselines &Metrics$^\dagger$& With label& Without label\\
        \midrule
        \multirow{4}{*}{\shortstack{CIFAR-10 }}
        &Leakage Rate ($\uparrow$) & 90.2\% & \textbf{99.4\%}\\
        &SSIM ($\uparrow$) & 0.618 & \textbf{0.703}\\
        &PSNR ($\uparrow$) & 20.566 & \textbf{21.867}\\  
        &LPIPS ($\downarrow$) & 0.409 & \textbf{0.358}\\   
        \hline
        \multirow{4}{*}{\shortstack{CIFAR-100}}
        &Leakage Rate ($\uparrow$) & 1.4\% & \textbf{98.6\%}\\
        &SSIM ($\uparrow$) & 0.199 & \textbf{0.663}\\
        &PSNR ($\uparrow$) & 14.244 & \textbf{21.514}\\  
        &LPIPS ($\downarrow$) & 0.604 & \textbf{0.371}\\ 
        \hline
        \multirow{4}{*}{MINI-ImageNet}
        &Leakage Rate ($\uparrow$) & 31.3\% &\textbf{ 89.8\%}\\
        &SSIM ($\uparrow$) & 0.424 & \textbf{0.635}\\
        &PSNR ($\uparrow$) & 18.617 &\textbf{23.850}\\  
        &LPIPS ($\downarrow$) & 0.569 & \textbf{0.412}\\   
        \hline
        \multirow{4}{*}{CelebA-Subset}
        &Leakage Rate ($\uparrow$) & 30.5\% & \textbf{100\%}\\
        &SSIM ($\uparrow$) & 0.472 &\textbf{0.794}\\
        &PSNR ($\uparrow$) & 15.206 & \textbf{28.290}\\  
        &LPIPS ($\downarrow$) & 0.541 & \textbf{0.292}\\   
       
        \bottomrule
    \end{tabular}
    \vspace{-0.3cm}
\end{table}



\textbf{Impact of label information in encoding.} 
In \sys, we introduce a novel label-agnostic image encoding scheme that separates the encoding process from local label information. We investigate the influence of label information on \sys, with the results presented in Table~\ref{tab:encoding_label}.
Notably, \sys without label information achieves a leakage rate of 89.8\% and high image quality on MINI-ImageNet, whereas \sys with label information achieves only a 31.3\% leakage rate with lower image quality. The results indicate that the reliance on label information limits performance, as the server must either possess or accurately estimate the local data distribution for an effective attack.

\subsection{Time Cost}
To evaluate the efficiency of \sys, we calculate its time cost, which consists of two components: training cost and memory cost. The training cost refers to the time required for the main task training of the local model, while the memory cost represents the time taken for training the secret model. 
The results are presented in Table~\ref{tab:train_cost} and \ref{tab:memory_cost}.  


These two tables show the time costs for training and memory tasks separately, reflecting their independence and respective dependencies. Training time is influenced by the local dataset size \( M \). With fixed training epochs and batch size, training time increases almost linearly with dataset size. For example, in the CIFAR-10 dataset, with \( M = 2000 \), training time is 16.4s, doubling to 32.8s when \( M \) is 4000.

In contrast, memory task time depends on the target theft quantity \( N \). Since we set the batch size equal to \( N \) for memory tasks, the time cost does not follow a strict linear relationship. For instance, with \( N = 512 \) in CIFAR-10, memory time is 19.2s.

In a practical theft scenario, specific configurations can make the task less noticeable. For instance, in CIFAR-10 with a local dataset size of 4000 and a theft of 512 images, training and memory times are 32.8s and 19.2s, respectively. The additional time for the memory task is close to half the original training time, making the theft less detectable.

\begin{table}[h]
\centering
\footnotesize
\setlength{\tabcolsep}{5pt}
\caption{Training Time Cost of Various Attack Components. \\
\textit{\(M\) denotes the local dataset size. All values are in seconds.}}
\begin{tabular}{l|ccc}
\toprule
Dataset & M=2000 & M=4000 & M=6000 \\ \midrule
CIFAR-10  & 16.4 & 32.8 & 48.3 \\
CIFAR-100 & 16.2 & 32.1 & 47.6 \\ 
ImageNet  & 21.5 & 43.7 & 63.2 \\
CelebA    & 20.2 & 41.8 & 60.9 \\ 
\bottomrule
\end{tabular}
\label{tab:train_cost}
\end{table}

\begin{table}[h]
\centering
\footnotesize
\setlength{\tabcolsep}{5pt}
\caption{Memory Time Cost of Various Attack Components. \\
\textit{\(N\) denotes the target theft quantity. All values are in seconds.}}
\begin{tabular}{l|ccc}
\toprule
Dataset & N=128 & N=256 & N=512 \\ \midrule
CIFAR-10  & 12.4 & 14.8 & 19.2 \\
CIFAR-100 & 11.6 & 15.7 & 20.8 \\ 
\midrule
Dataset & N=32 & N=48 & N=64 \\ \midrule
ImageNet  & 22.1 & 24.3 & 26.8 \\
CelebA    & 18.2 & 19.7 & 21.2 \\ 
\bottomrule
\end{tabular}
\label{tab:memory_cost}
\vspace{-0.3cm}
\end{table}

\begin{table*}[tt]

\caption{{\color{black}Robustness of \sys to various advanced defenses.} }
	\label{tab:com1}
	\centering
	\setlength\tabcolsep{3pt}
         \footnotesize
	\begin{tabular}{cl|cccccccccc}
		\toprule
       \multirow{2}{*}{\shortstack{Dataset}}&\multirow{2}{*}{\shortstack{Metrics$^\dagger$}} & \multicolumn{2}{c|}{Gradient Pruning} & \multicolumn{2}{c|}{Gradient Clipping}  & \multicolumn{2}{c|}{Noise Perturbation} & \multirow{2}{*}{\shortstack{D-SNR Detection}}\\
       & & $\tau$= 1e{-}6 & \multicolumn{1}{c|}{$\tau$= 1e{-}5} & max\_norm = 5 & \multicolumn{1}{c|}{max\_norm = 1} & $\epsilon$= 1e{-}3 & \multicolumn{1}{c|}{$\epsilon$= 4e{-}3} & \\
   	\midrule

          \multirow{4}{*}{\shortstack{CIFAR-10}}
          &Leakage ($\uparrow$) & 511.0 & 448.0 & 511.0 & 511.0 & 511.8 & 431.4 & 512.0 \\
          &SSIM ($\uparrow$) & 0.746 & 0.590 & 0.748 & 0.748 & 0.760 & 0.593 & 0.788 \\
          &PSNR ($\uparrow$) & 22.637 & 20.186 & 22.692 & 22.675 & 22.537 & 19.043 & 23.687 \\  
          &LPIPS ($\downarrow$) & 0.325 & 0.428 & 0.324 & 0.323 & 0.317 & 0.427 & 0.264 \\   
          \hline
          \multirow{4}{*}{\shortstack{CIFAR-100}}
          &Leakage ($\uparrow$) & 510.0 & 387.0 & 511.0 & 511.0 & 505.8 & 502.8 & 506.4 \\
          &SSIM ($\uparrow$) & 0.699 & 0.568 & 0.712 & 0.724 & 0.689 & 0.685 & 0.712 \\
          &PSNR ($\uparrow$) & 22.080 & 20.248 & 22.304 & 22.523 & 21.966 & 21.899 & 22.486 \\  
          &LPIPS ($\downarrow$) & 0.345 & 0.427 & 0.335 & 0.325 & 0.349 & 0.352 & 0.183 \\   
          \hline
          \multirow{4}{*}{\shortstack{MINI-ImageNet}}
          &Leakage ($\uparrow$) & 32.0 & 32.0 & 32.0 & 31.0 & 32.0 & 11.4 & 32.0 \\
          &SSIM ($\uparrow$) & 0.829 & 0.823 & 0.830 & 0.779 & 0.735 & 0.485 & 0.805\\
          &PSNR ($\uparrow$) & 28.361 & 28.157 & 28.373 & 26.845 & 26.039 & 21.356 & 27.311 \\  
          &LPIPS ($\downarrow$) & 0.222 & 0.230 & 0.222 & 0.281 & 0.331 & 0.520 & 0.248 \\      
          \hline
          \multirow{4}{*}{\shortstack{CelebA Subset}}
          &Leakage ($\uparrow$) & 32.0 & 32.0 & 32.0 & 32.0 & 32.0 & 13.8 & 32.0 \\
          &SSIM ($\uparrow$) & 0.907 & 0.903 & 0.914 & 0.919 & 0.849 & 0.489 & 0.911 \\
          &PSNR ($\uparrow$) & 32.659 & 32.486 & 33.085 & 33.421 & 30.744 & 22.423 & 33.041 \\  
          &LPIPS ($\downarrow$) & 0.127 & 0.131 & 0.117 & 0.110 & 0.231 & 0.557 & 0.123\\   
         
		\bottomrule
	\end{tabular}
\end{table*}

\section{Robustness to State-of-the-art Defenses}

In this section, we investigate the effectiveness of \sys under state-of-the-art data reconstruction defenses, including D-SNR \cite{garov2023hiding}, noise perturbation, gradient pruning, and gradient clipping. Additionally, we monitor loss changes as a defense mechanism to assess \sys's resilience against detection. 

\subsection{D-SNR Detection}


Disaggregation signal-to-noise ratio (D-SNR) is a novel metric for detecting vulnerabilities in federated learning, specifically against disaggregation attacks. D-SNR signals a potential attack when the gradient of a single example outweighs the batch gradient, suggesting an adversary has isolated an individual example from the batch. 
This metric is essential for evaluating the risk of data leakage by analyzing gradients, bypassing the need for costly optimization-based attack simulations. By ensuring that any layer with potential disaggregation has a high D-SNR, clients can detect and skip vulnerable training rounds.

When applying D-SNR, we observe that \sys can still succeed by avoiding dominant gradients within any single training round. 
The success of \sys lies in its multi-round, incremental data embedding approach, which encodes small portions of sensitive data across multiple rounds.
This strategy distributes the data over several rounds and merges gradients with shared parameters, keeping gradient magnitudes low enough to evade D-SNR detection.
Moreover, \sys’s sparse encoding minimizes gradient impact per round, effectively bypassing the D-SNR threshold.



\subsection{Noise Perturbation}


Defenders can enhance a model's robustness by adding noise to the gradients, which involves applying small, continuous perturbations. A common method is to introduce Gaussian noise after each backpropagation step, referred to as the diffusion term \cite{liu2020does}. This technique generates random variations in gradient values, thereby mitigating the risk of data leakage during updates.

To assess the impact of noise-based defenses, we applied Gaussian noise directly to the gradients in our experiments, with noise levels set at \(1 \times 10^{-3}\) and \(4 \times 10^{-3}\). Despite these relatively high noise levels, the results indicate that \texttt{\sys} remains effective. When the noise is set to \(1 \times 10^{-3}\), the effectiveness of data theft is nearly unaffected. With an increased noise level of \(4 \times 10^{-3}\), \texttt{\sys} still achieves a leakage of 431.4 on CIFAR-10 with a target theft of 512 images, and for CelebA, with a target of 32 images, \texttt{\sys} successfully steals 13.8 images. 
\sys succeeds due to its low-magnitude, multi-round embedding strategy, which enables it to accumulate data across multiple rounds while maintaining effectiveness even with added noise.

\subsection{Gradient Pruning}
Gradient pruning \cite{zhang2019gradient} reduces computational complexity in neural networks by selectively pruning channels based on the importance of their gradients. It evaluates the significance of each channel using the mean gradient of the feature maps, pruning those with lower mean gradients that have less impact on the loss function. It can decrease the network's size and floating-point operations (FLOPs). 

In the experiments, we selected two threshold values for gradient pruning, \(\tau = 1 \times 10^{-6}\) and \(\tau = 1 \times 10^{-5}\), to control the level of pruning applied. The threshold \(\tau\) determines the minimum mean gradient magnitude required for a channel to be retained. Lower values of \(\tau\) (e.g., \(1 \times 10^{-6}\)) result in fewer channels being pruned, preserving more information, while higher values (e.g., \(1 \times 10^{-5}\)) increase pruning aggressiveness, removing more channels with lower gradient contributions. Despite the use of gradient pruning, \sys remains effective due to its multi-round, low-intensity embedding strategy. Our method does not rely on high-gradient channels in any single round but instead distributes data across multiple rounds and channels, with each round embedding minimal information. 
By leveraging sparse encoding and parameter sharing, \sys disperses data across numerous channels, allowing it to circumvent pruning. Even if some channels are pruned, the cumulative effect of the attack remains intact, ensuring successful data leakage.

\subsection{Gradient Clipping}
Gradient clipping \cite{liu2019channel} is a defense technique that limits the norm of gradients during backpropagation to prevent excessively large updates that could destabilize training, especially when dealing with exploding gradients. When the gradient norm exceeds a predefined threshold, it is rescaled to match the threshold. This helps maintain training stability by controlling the gradient magnitude, allowing the model to navigate non-smooth regions of the loss landscape more effectively, leading to faster convergence and potentially improved generalization.
In the experiments, we applied gradient clipping with two threshold values, \( \text{max\_norm} = 1 \) and \( \text{max\_norm} = 5 \), to control the magnitude of the gradients during backpropagation. These thresholds help limit the gradient updates, with \( \text{max\_norm} = 1 \) enforcing stricter clipping and \( \text{max\_norm} = 5 \) allowing slightly larger updates. By constraining the gradient norm, we aim to prevent excessively large updates that could destabilize training or signal potential data leakage.

However, we can see that \sys remains effective under gradient clipping because it does not rely on large gradient values in a single round. Instead, it employs a multi-round, low-magnitude embedding strategy, where each round only contributes a small amount of data, keeping the gradient norm within the clipping limits. 

\subsection{Loss Change Monitor}
During model training, clients may monitor the change in loss to detect anomalies. To prevent detection, the loss should appear normal and remain within expected fluctuations throughout training. The loss change is shown in Figure~\ref{fig:loss} (appendix). We can see that there is a slight increase in the first round when the client trains the local model alongside the memory task. However, the loss quickly stabilizes and gradually decreases, resembling the pattern of standard training, thereby reducing the risk of drawing the client's attention. In normal training, minor increases in loss can occur due to factors such as model adjustments and data variability. Thus, this subtle rise is unlikely to raise suspicion, as it resembles natural training fluctuations, demonstrating the evasiveness of \sys.

\section{Discussion}
\subsection{Secure Aggregation} \label{secureagg}
Secure aggregation presents a significant challenge for data reconstruction attacks, as it only allows the server to access aggregated updates from all clients, blocking access to individual updates \cite{bonawitz2017practical}. Many attacks rely on reconstructing high-quality images from individual client updates, which makes them largely ineffective when secure aggregation is in place \cite{wen2022fishing,boenisch2023curious,zhang2022compromise}.

Federated Learning (FL) also faces communication bottlenecks due to the limited bandwidth of edge devices and their resource constraints \cite{pfeiffer2024aggregating,hsieh2017gaia,chen2021communication}. To address this, methods like reducing the number of clients per round \cite{mcmahan2017communication,luo2022tackling} and applying gradient compression techniques (e.g., quantization, sparsification, low-rank approximation) \cite{nader2021adaptive,wen2022federated,chen2022fedobd} have been developed. Gradient sparsification is especially useful, as it filters out less important gradients, reducing the amount of data transmitted during updates.

We discovered that by combining our approach with communication acceleration strategies, it is possible to bypass secure aggregation. These strategies can create model inconsistencies across clients due to differences in data distribution and gradient sparsification. Attackers can exploit these inconsistencies to extract gradient information. For instance, in APF \cite{chen2021communication}, a global frozen mask is used to synchronize parameters across clients. By manipulating this mask with malicious code (e.g., setting it to zero or inverting it), attackers can expose the target user’s gradient information.

\subsection{Generalize \sys to NLP}
Although \sys was originally designed for applications in computer vision, its principles can be adapted to natural language processing (NLP) tasks with slight modifications. In NLP, the secret model can be embedded within transformers or similar architectures by selecting and masking specific layers or attention heads. Instead of handling images, the model memorizes token sequences, using techniques like tokenization for indexing to distinguish between different samples.

To process longer text sequences, block partitioning can be applied by splitting the text into smaller segments. These segments are treated as individual inputs for memorization, with token-level information used to reassemble them during reconstruction. The loss function would focus on minimizing the difference between predicted and actual token sequences, allowing the model to covertly learn and store sensitive text data. In the future, we will explore the effectiveness and scalability of \sys in NLP tasks.

\subsection{Potential Countermeasures}
To defend against the covert data-reconstruction mechanisms in \sys, two potential countermeasures can be applied.

First, regularly verifying the integrity of model parameters before and after training is crucial. Techniques such as hash-based verification can help detect unauthorized modifications by comparing model snapshots across different training rounds. This approach can flag suspicious changes that might suggest data-stealing activities early on. However, it may not be effective if the \sys introduces subtle changes that blend with normal updates. These changes can be distributed across numerous parameters, making them difficult to detect through simple integrity checks. Specifically, malicious code injections can modify the model in ways that appear innocuous when viewed in isolation but collectively enable a stealthy attack.

Second, gradient noise injection techniques, such as differential privacy, can make it more difficult for attackers to recover sensitive data from model updates. By adding controlled noise to the gradients during training, the accuracy of extracted data is reduced, limiting \sys's ability to reconstruct sensitive token sequences. However, excessive noise may degrade the model's performance, reducing its utility for legitimate users, making it challenging to find a balance between security and accuracy.

In the future, designing more effective defenses against \sys will be necessary to mitigate the risks of such attacks.

\section{Conclusion}
In this paper, we introduced a novel data reconstruction attack against federated learning. Unlike traditional methods that rely on conspicuous changes to architecture or parameters, \sys injects malicious code during training, enabling undetected data theft. By covertly training a hidden model through parameter sharing, \sys efficiently extracts private data. To improve performance, we proposed a Fibonacci-based indexing and a block partitioning strategy that enhances the attack's ability to handle high-resolution datasets and large batch sizes.
Extensive experiments show that \sys can bypass state-of-the-art detection methods while effectively handling high-resolution datasets and large-scale theft scenarios.

\bibliographystyle{plain}
\bibliography{plain}

\begin{thebibliography}{10}

\bibitem{abadi2016deep}
Martin Abadi, Andy Chu, Ian Goodfellow, H~Brendan McMahan, Ilya Mironov, Kunal Talwar, and Li~Zhang.
\newblock Deep learning with differential privacy.
\newblock In {\em SIGSAC Conference on Computer and Communications Security}, pages 308--318, 2016.

\bibitem{amit2023transpose}
Guy Amit, Mosh Levy, and Yisroel Mirsky.
\newblock Transpose attack: Stealing datasets with bidirectional training.
\newblock {\em arXiv preprint arXiv:2311.07389}, 2023.

\bibitem{aono2017privacy}
Yoshinori Aono, Takuya Hayashi, Lihua Wang, Shiho Moriai, et~al.
\newblock Privacy-preserving deep learning via additively homomorphic encryption.
\newblock {\em IEEE transactions on information forensics and security}, 13(5):1333--1345, 2017.

\bibitem{apostolico1987robust}
Alberto Apostolico and A~Fraenkel.
\newblock Robust transmission of unbounded strings using fibonacci representations.
\newblock {\em IEEE Transactions on Information Theory}, 33(2):238--245, 1987.

\bibitem{bagdasaryan2021blind}
Eugene Bagdasaryan and Vitaly Shmatikov.
\newblock Blind backdoors in deep learning models.
\newblock In {\em USENIX Security Symposium}, pages 1505--1521, 2021.

\bibitem{boenisch2023reconstructing}
Franziska Boenisch, Adam Dziedzic, Roei Schuster, Ali~Shahin Shamsabadi, Ilia Shumailov, and Nicolas Papernot.
\newblock Reconstructing individual data points in federated learning hardened with differential privacy and secure aggregation.
\newblock In {\em European Symposium on Security and Privacy}, pages 241--257. IEEE, 2023.

\bibitem{boenisch2023curious}
Franziska Boenisch, Adam Dziedzic, Roei Schuster, Ali~Shahin Shamsabadi, Ilia Shumailov, and Nicolas Papernot.
\newblock When the curious abandon honesty: Federated learning is not private.
\newblock In {\em IEEE European Symposium on Security and Privacy}, pages 175--199, 2023.

\bibitem{bonawitz2017practical}
Keith Bonawitz, Vladimir Ivanov, Ben Kreuter, Antonio Marcedone, H~Brendan McMahan, Sarvar Patel, Daniel Ramage, Aaron Segal, and Karn Seth.
\newblock Practical secure aggregation for privacy-preserving machine learning.
\newblock In {\em SIGSAC Conference on Computer and Communications Security}, pages 1175--1191. ACM, 2017.

\bibitem{chen2021understanding}
Cangxiong Chen and Neill~DF Campbell.
\newblock Understanding training-data leakage from gradients in neural networks for image classification.
\newblock {\em arXiv preprint arXiv:2111.10178}, 2021.

\bibitem{chen2021communication}
Chen Chen, Hong Xu, Wei Wang, Baochun Li, Bo~Li, Li~Chen, and Gong Zhang.
\newblock Communication-efficient federated learning with adaptive parameter freezing.
\newblock In {\em IEEE International Conference on Distributed Computing Systems}, pages 1--11, 2021.

\bibitem{chen2014qos}
Yanjiao Chen, Kaishun Wu, and Qian Zhang.
\newblock {From QoS to QoE: A} tutorial on video quality assessment.
\newblock {\em IEEE Communications Surveys \& Tutorials}, 17(2):1126--1165, 2014.

\bibitem{chen2022fedobd}
Yuanyuan Chen, Zichen Chen, Pengcheng Wu, and Han Yu.
\newblock Fedobd: Opportunistic block dropout for efficiently training large-scale neural networks through federated learning.
\newblock {\em arXiv preprint arXiv:2208.05174}, 2022.

\bibitem{duan2020towards}
Ruian Duan, Omar Alrawi, Ranjita~Pai Kasturi, Ryan Elder, Brendan Saltaformaggio, and Wenke Lee.
\newblock Towards measuring supply chain attacks on package managers for interpreted languages.
\newblock {\em arXiv preprint arXiv:2002.01139}, 2020.

\bibitem{imagentte}
fast.ai.
\newblock Imagenette: A smaller subset of 10 easily classified classes from imagenet.
\newblock {\em Available: https://github.com/fastai/imagenette}.

\bibitem{fowl2021robbing}
Liam Fowl, Jonas Geiping, Wojtek Czaja, Micah Goldblum, and Tom Goldstein.
\newblock Robbing the fed: Directly obtaining private data in federated learning with modified models.
\newblock {\em arXiv preprint arXiv:2110.13057}, 2021.

\bibitem{fowl2022decepticons}
Liam Fowl, Jonas Geiping, Steven Reich, Yuxin Wen, Wojtek Czaja, Micah Goldblum, and Tom Goldstein.
\newblock Decepticons: Corrupted transformers breach privacy in federated learning for language models.
\newblock {\em arXiv preprint arXiv:2201.12675}, 2022.

\bibitem{garov2023hiding}
Kostadin Garov, Dimitar~I Dimitrov, Nikola Jovanovi{\'c}, and Martin Vechev.
\newblock Hiding in plain sight: Disguising data stealing attacks in federated learning.
\newblock {\em arXiv preprint arXiv:2306.03013}, 2023.

\bibitem{geiping2020inverting}
Jonas Geiping, Hartmut Bauermeister, Hannah Dr{\"o}ge, and Michael Moeller.
\newblock Inverting gradients-how easy is it to break privacy in federated learning?
\newblock {\em Advances in Neural Information Processing Systems}, 33:16937--16947, 2020.

\bibitem{geyer2017differentially}
Robin~C Geyer, Tassilo Klein, and Moin Nabi.
\newblock Differentially private federated learning: A client level perspective.
\newblock {\em arXiv preprint arXiv:1712.07557}, 2017.

\bibitem{gupta2022recovering}
Samyak Gupta, Yangsibo Huang, Zexuan Zhong, Tianyu Gao, Kai Li, and Danqi Chen.
\newblock Recovering private text in federated learning of language models.
\newblock {\em Advances in Neural Information Processing Systems}, 35:8130--8143, 2022.

\bibitem{howard2020fastai}
Jeremy Howard and Sylvain Gugger.
\newblock Fastai: A layered api for deep learning.
\newblock {\em Information}, 11(2):108, 2020.

\bibitem{hsieh2017gaia}
Kevin Hsieh, Aaron Harlap, Nandita Vijaykumar, Dimitris Konomis, Gregory~R Ganger, Phillip~B Gibbons, and Onur Mutlu.
\newblock Gaia:$\{$Geo-Distributed$\}$ machine learning approaching $\{$LAN$\}$ speeds.
\newblock In {\em USENIX Symposium on Networked Systems Design and Implementation}, pages 629--647, 2017.

\bibitem{jeon2021gradient}
Jinwoo Jeon, Kangwook Lee, Sewoong Oh, Jungseul Ok, et~al.
\newblock Gradient inversion with generative image prior.
\newblock {\em Advances in Neural Information Processing Systems}, 34:29898--29908, 2021.

\bibitem{kairouz2021advances}
Peter Kairouz, H~Brendan McMahan, Brendan Avent, Aur{\'e}lien Bellet, Mehdi Bennis, Arjun~Nitin Bhagoji, Kallista Bonawitz, Zachary Charles, Graham Cormode, Rachel Cummings, et~al.
\newblock Advances and open problems in federated learning.
\newblock {\em Foundations and trends{\textregistered} in machine learning}, 14(1--2):1--210, 2021.

\bibitem{konecny2016federated}
Jakub Konecn{\`y}, H~Brendan McMahan, Felix~X Yu, Peter Richt{\'a}rik, Ananda~Theertha Suresh, and Dave Bacon.
\newblock Federated learning: Strategies for improving communication efficiency.
\newblock {\em arXiv preprint arXiv:1610.05492}, 8, 2016.

\bibitem{krizhevsky2009learning}
Alex Krizhevsky and Geoffrey Hinton.
\newblock Learning multiple layers of features from tiny images.
\newblock 2009.

\bibitem{lam2021gradient}
Maximilian Lam, Gu-Yeon Wei, David Brooks, Vijay~Janapa Reddi, and Michael Mitzenmacher.
\newblock Gradient disaggregation: Breaking privacy in federated learning by reconstructing the user participant matrix.
\newblock In {\em International Conference on Machine Learning}, pages 5959--5968. PMLR, 2021.

\bibitem{li2022auditing}
Zhuohang Li, Jiaxin Zhang, Luyang Liu, and Jian Liu.
\newblock Auditing privacy defenses in federated learning via generative gradient leakage.
\newblock In {\em IEEE/CVF Conference on Computer Vision and Pattern Recognition}, pages 10132--10142, 2022.

\bibitem{liu2019channel}
Congcong Liu and Huaming Wu.
\newblock Channel pruning based on mean gradient for accelerating convolutional neural networks.
\newblock {\em Signal Processing}, 156:84--91, 2019.

\bibitem{liu2020does}
Xuanqing Liu, Tesi Xiao, Si~Si, Qin Cao, Sanjiv Kumar, and Cho-Jui Hsieh.
\newblock How does noise help robustness? explanation and exploration under the neural sde framework.
\newblock In {\em IEEE/CVF Conference on Computer Vision and Pattern Recognition}, pages 282--290, 2020.

\bibitem{liu2015deep}
Ziwei Liu, Ping Luo, Xiaogang Wang, and Xiaoou Tang.
\newblock Deep learning face attributes in the wild.
\newblock In {\em IEEE International Conference on Computer Vision}, pages 3730--3738, 2015.

\bibitem{lu2021discriminator}
Shaohao Lu, Yuqiao Xian, Ke~Yan, Yi~Hu, Xing Sun, Xiaowei Guo, Feiyue Huang, and Wei-Shi Zheng.
\newblock Discriminator-free generative adversarial attack.
\newblock In {\em ACM International Conference on Multimedia}, pages 1544--1552, 2021.

\bibitem{luo2022tackling}
Bing Luo, Wenli Xiao, Shiqiang Wang, Jianwei Huang, and Leandros Tassiulas.
\newblock Tackling system and statistical heterogeneity for federated learning with adaptive client sampling.
\newblock In {\em IEEE INFOCOM Conference on Computer Communications}, pages 1739--1748, 2022.

\bibitem{luo2021feature}
Xinjian Luo, Yuncheng Wu, Xiaokui Xiao, and Beng~Chin Ooi.
\newblock Feature inference attack on model predictions in vertical federated learning.
\newblock In {\em International Conference on Data Engineering}, pages 181--192. IEEE, 2021.

\bibitem{mcmahan2017communication}
Brendan McMahan, Eider Moore, Daniel Ramage, Seth Hampson, and Blaise~Aguera y~Arcas.
\newblock Communication-efficient learning of deep networks from decentralized data.
\newblock In {\em Artificial Intelligence and Statistics}, pages 1273--1282. PMLR, 2017.

\bibitem{melis2019exploiting}
Luca Melis, Congzheng Song, Emiliano De~Cristofaro, and Vitaly Shmatikov.
\newblock Exploiting unintended feature leakage in collaborative learning.
\newblock In {\em IEEE Symposium on Security and Privacy (SP)}, pages 691--706, 2019.

\bibitem{nader2021adaptive}
Bouacida Nader, Hou Jiahui, Zang Hui, and Liu Xin.
\newblock Adaptive federated dropout: Improving communication efficiency and generalization for federated learning.
\newblock In {\em IEEE Conference on Computer Communications Workshops, Vancouver, BC, Canada}, pages 10--13, 2021.

\bibitem{nasr2019comprehensive}
Milad Nasr, Reza Shokri, and Amir Houmansadr.
\newblock Comprehensive privacy analysis of deep learning: Passive and active white-box inference attacks against centralized and federated learning.
\newblock In {\em IEEE Symposium on Security and Privacy}, pages 739--753, 2019.

\bibitem{ott2019fairseq}
M~Ott.
\newblock fairseq: A fast, extensible toolkit for sequence modeling.
\newblock {\em arXiv preprint arXiv:1904.01038}, 2019.

\bibitem{pasquini2022eluding}
Dario Pasquini, Danilo Francati, and Giuseppe Ateniese.
\newblock Eluding secure aggregation in federated learning via model inconsistency.
\newblock In {\em ACM SIGSAC Conference on Computer and Communications Security}, pages 2429--2443, 2022.

\bibitem{pfeiffer2024aggregating}
Kilian Pfeiffer, Ramin Khalili, and J{\"o}rg Henkel.
\newblock Aggregating capacity in fl through successive layer training for computationally-constrained devices.
\newblock {\em Advances in Neural Information Processing Systems}, 36, 2024.

\bibitem{phong2017privacy}
Le~Trieu Phong, Yoshinori Aono, Takuya Hayashi, Lihua Wang, and Shiho Moriai.
\newblock Privacy-preserving deep learning: Revisited and enhanced.
\newblock In {\em Applications and Techniques in Information Security: 8th International Conference}, pages 100--110. Springer, 2017.

\bibitem{sun2021soteria}
Jingwei Sun, Ang Li, Binghui Wang, Huanrui Yang, Hai Li, and Yiran Chen.
\newblock Soteria: Provable defense against privacy leakage in federated learning from representation perspective.
\newblock In {\em IEEE/CVF Conference on Computer Vision and Pattern Recognition}, pages 9311--9319, 2021.

\bibitem{wang2020sapag}
Yijue Wang, Jieren Deng, Dan Guo, Chenghong Wang, Xianrui Meng, Hang Liu, Caiwen Ding, and Sanguthevar Rajasekaran.
\newblock Sapag: A self-adaptive privacy attack from gradients.
\newblock {\em arXiv preprint arXiv:2009.06228}, 2020.

\bibitem{wang2019beyond}
Zhibo Wang, Mengkai Song, Zhifei Zhang, Yang Song, Qian Wang, and Hairong Qi.
\newblock Beyond inferring class representatives: User-level privacy leakage from federated learning.
\newblock In {\em IEEE INFOCOM Conference on Computer Communications}, pages 2512--2520, 2019.

\bibitem{wei2020framework}
Wenqi Wei, Ling Liu, Margaret Loper, Ka-Ho Chow, Mehmet~Emre Gursoy, Stacey Truex, and Yanzhao Wu.
\newblock A framework for evaluating gradient leakage attacks in federated learning.
\newblock {\em arXiv preprint arXiv:2004.10397}, 2020.

\bibitem{wei2021gradient}
Wenqi Wei, Ling Liu, Yanzhao Wu, Gong Su, and Arun Iyengar.
\newblock Gradient-leakage resilient federated learning.
\newblock In {\em International Conference on Distributed Computing Systems}, pages 797--807. IEEE, 2021.

\bibitem{wen2022federated}
Dingzhu Wen, Ki-Jun Jeon, and Kaibin Huang.
\newblock Federated dropout—a simple approach for enabling federated learning on resource constrained devices.
\newblock {\em IEEE Wireless Communications Letters}, 11(5):923--927, 2022.

\bibitem{wen2022fishing}
Yuxin Wen, Jonas Geiping, Liam Fowl, Micah Goldblum, and Tom Goldstein.
\newblock Fishing for user data in large-batch federated learning via gradient magnification.
\newblock {\em arXiv preprint arXiv:2202.00580}, 2022.

\bibitem{wolf2019huggingface}
T~Wolf.
\newblock Huggingface's transformers: State-of-the-art natural language processing.
\newblock {\em arXiv preprint arXiv:1910.03771}, 2019.

\bibitem{xiang2021patchguard}
Chong Xiang, Arjun~Nitin Bhagoji, Vikash Sehwag, and Prateek Mittal.
\newblock Patchguard: A provably robust defense against adversarial patches via small receptive fields and masking.
\newblock In {\em 30th USENIX Security Symposium}, 2021.

\bibitem{yang2022using}
Haomiao Yang, Mengyu Ge, Kunlan Xiang, and Jingwei Li.
\newblock Using highly compressed gradients in federated learning for data reconstruction attacks.
\newblock {\em IEEE Transactions on Information Forensics and Security}, 18:818--830, 2022.

\bibitem{zeckendorf1972representations}
{\'E}douard Zeckendorf.
\newblock Representations des nombres naturels par une somme de nombres de fibonacci on de nombres de lucas.
\newblock {\em Bulletin de La Society Royale des Sciences de Liege}, pages 179--182, 1972.

\bibitem{zhang2019gradient}
Jingzhao Zhang, Tianxing He, Suvrit Sra, and Ali Jadbabaie.
\newblock Why gradient clipping accelerates training: A theoretical justification for adaptivity.
\newblock {\em arXiv preprint arXiv:1905.11881}, 2019.

\bibitem{zhang2018unreasonable}
Richard Zhang, Phillip Isola, Alexei~A Efros, Eli Shechtman, and Oliver Wang.
\newblock The unreasonable effectiveness of deep features as a perceptual metric.
\newblock In {\em IEEE Conference on Computer Vision and Pattern Recognition}, pages 586--595, 2018.

\bibitem{zhang2022compromise}
Shuaishuai Zhang, Jie Huang, Zeping Zhang, and Chunyang Qi.
\newblock Compromise privacy in large-batch federated learning via malicious model parameters.
\newblock In {\em International Conference on Algorithms and Architectures for Parallel Processing}, pages 63--80. Springer, 2022.

\bibitem{zhao2023resource}
Joshua~C Zhao, Ahmed~Roushdy Elkordy, Atul Sharma, Yahya~H Ezzeldin, Salman Avestimehr, and Saurabh Bagchi.
\newblock The resource problem of using linear layer leakage attack in federated learning.
\newblock In {\em IEEE/CVF Conference on Computer Vision and Pattern Recognition}, pages 3974--3983, 2023.

\bibitem{zhao2023loki}
Joshua~Christian Zhao, Atul Sharma, Ahmed~Roushdy Elkordy, Yahya~H Ezzeldin, Salman Avestimehr, and Saurabh Bagchi.
\newblock Loki: Large-scale data reconstruction attack against federated learning through model manipulation.
\newblock In {\em IEEE Symposium on Security and Privacy}, pages 30--30. IEEE Computer Society, 2023.

\bibitem{zhu2020r}
Junyi Zhu and Matthew Blaschko.
\newblock R-gap: Recursive gradient attack on privacy.
\newblock {\em arXiv preprint arXiv:2010.07733}, 2020.

\bibitem{zhu2019deep}
Ligeng Zhu, Zhijian Liu, and Song Han.
\newblock Deep leakage from gradients.
\newblock {\em Advances in Neural Information Processing Systems}, 32, 2019.

\end{thebibliography}

\appendix

\subsection*{A. Datasets and Models}

\emph{CIFAR-10}. CIFAR-10 \cite{krizhevsky2009learning} contains 60,000 images belonging to 10 classes. Each sample has a dimension of 32 $\times$ 32. We randomly select 50,000 samples as the training set, and the remaining 10,000 samples as the test set. In the experiments, we employ the ResNet-18 architecture to train the victim model. The local training process spans 10 epochs, with a learning rate initially set to 0.1. We use the SGD optimizer with a weight decay of 0.001 to prevent overfitting. To refine training over time, we apply a learning rate scheduler that decays the rate by a factor of 0.9 every epoch. The training batch size is set to 32. 

\emph{CIFAR-100}. CIFAR-100 dataset \cite{krizhevsky2009learning} closely resembles CIFAR-10, differing in the number of classes. It comprises 100 classes, each containing 600 images. The dataset is split into 500 training images and 100 testing images per class, amounting to a comprehensive set of diverse visual data for classification tasks. In the experiments, we employ the ResNet-18 architecture to train the victim model. The local training details for the CIFAR-100 dataset are consistent with those of CIFAR-10.

\emph{MINI-ImageNet.}  MINI-ImageNet\cite{imagentte} is a subset of ImageNet, widely used in the research community  \cite{xiang2021patchguard,lu2021discriminator}.  It consists of 100 classes, each containing 600 images. For our experiments, we selected 40,000 images from the dataset, with 400 images per class across 100 classes, to ensure balanced representation across categories. Each image has a high resolution with a dimension of 224 $\times$ 224. In the experiments, we train a ResNet-18 model for 10 epochs as the victim model. The local training details for the MINI-ImageNet dataset are consistent with those of CIFAR-10.

\emph{CelebA.} CelebA \cite{liu2015deep} is a large-scale face attributes dataset containing 10,177 identities with 202,599 face images. Each image includes annotations for 5 landmark locations and 40 binary attributes, and has a high resolution of 224 × 224 pixels. In our experiments, we selected 1,000 identities from the dataset, with 20 images per identity, totaling 20,000 images. For local training, we kept the training parameters consistent with those used for CIFAR-10.

\subsection*{B. Evaluation Metrics}

\textbf{Leakage:} The leakage quantifies the number of images that can be extracted from the total dataset in the given attack rounds.

\textbf{Leakage Rate:} The leakage rate quantifies the proportion of images that can be extracted from the total dataset in the given attack rounds.
\begin{equation}
\text{Leakage Rate} = \frac{E}{N},
\end{equation}
where $E$ is the number of extracted images, and $N$ is the total number of images in the target dataset.

\textbf{SSIM}. SSIM is a commonly-used Quality-of-Experience (QoE) metric \cite{chen2014qos} that quantifies the differences in luminance, contrast, and structure between the original image and the distorted image.
\begin{equation}
SSIM = A(x,x')^{\alpha} B(x,x')^{\beta} C(x,x')^{\gamma},
\end{equation}
where $A(x,x'), B(x,x')$, and $C(x,x')$ quantify the luminance similarity, contrast similarity, and structure similarity between the original image $x$ and the distorted image $x'$. $\alpha, \beta$, and $\gamma$ are parameters in the range $[0, 1]$. 

\textbf{PSNR}. PSNR is computed based on MSE (Mean Squared Error) regarding the signal energy. 
\begin{equation}
\begin{split}
    PSNR = 10\log_{10} \frac{E}{MSE},\\
    MSE = \frac{1}{N} \sum_i (x_i' - x_i)^2,
\end{split}
\end{equation}
where $E$ is the maximum signal energy.

\textbf{LPIPS}. LPIPS \cite{zhang2018unreasonable} measures the similarity between two images based on the idea that the human visual system processes images in a hierarchical manner, where lower-level features, e.g., edges and textures, are processed before higher-level features, e.g., objects and scenes. The LPIPS metric uses a deep neural network to calculate the similarity between the two images. 
\begin{equation}
    LPIPS(A,B) = \sum_i w_i * ||F_i(A) - F_i(B)||^2
\end{equation}
where $F_i(A)$ and $F_i(B)$ are the feature representations of images $A$ and $B$ at layer $i$ of the pre-trained neural network, $||.||$ denotes the L$_2$ norm, and $w_i$ is a weight that controls the relative importance of each layer. 
The smaller the value, the more similar the two images are.

\begin{figure}[tt]
    \centering
    \includegraphics[width=0.35\textwidth]{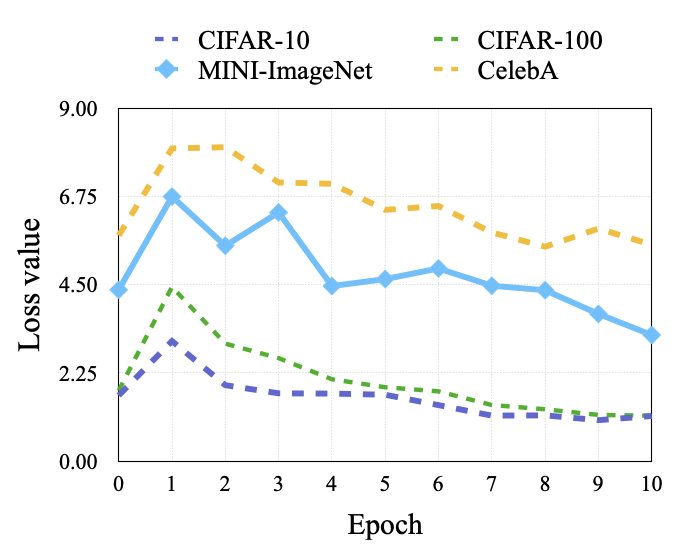}
    \caption{{\color{black}Impact of the loss change.}} 
    \label{fig:loss}
\end{figure}

\begin{table}[t]
    \caption{Impact of different index design methods on attack performance.} 
    \label{tab:encoding}
    \centering
    \footnotesize
    \begin{tabular}{lc|cccc}
        \toprule
        Baselines &Metrics$^\dagger$&Binary& Gray & Fibonacci\\
        \midrule
        \multirow{4}{*}{\shortstack{CIFAR-10 }}
        &Leakage Rate ($\uparrow$) & 84.4\% & 77.3\% & 85.9\%\\
        &SSIM ($\uparrow$) & 0.604 & 0.567 & 0.607\\
        &PSNR ($\uparrow$) & 28.037 & 26.736 &28.105\\  
        &LPIPS ($\downarrow$) &0.401 & 0.425 & 0.401\\   
        \hline
        \multirow{4}{*}{\shortstack{CIFAR-100}}
        &Leakage Rate ($\uparrow$) & 83.6\% & 79.7\% & 84.4\%\\
        &SSIM ($\uparrow$) &0.829& 0.781 & 0.827\\
        &PSNR ($\uparrow$) &32.918 & 29.873 & 32.142\\  
        &LPIPS ($\downarrow$) & 0.143 & 0.130& 0.147\\ 
        \hline
        \multirow{4}{*}{MINI-ImageNet}
        &Leakage Rate ($\uparrow$) & 77.8\% & 80.0\% &84.4\%\\
        &SSIM ($\uparrow$) & 0.598 & 0.572 & 0.612\\
        &PSNR ($\uparrow$) & 23.179 & 22.239 &23.535\\  
        &LPIPS ($\downarrow$) & 0.464 &0.490&0.451\\   
        \hline
        \multirow{4}{*}{CelebA-Subset}
        &Leakage Rate ($\uparrow$) &93.8\%& 81.3\% &93.8\%\\
        &SSIM ($\uparrow$) & 0.631 &0.736& 0.661\\
        &PSNR ($\uparrow$) & 21.565 &25.307& 22.667\\  
        &LPIPS ($\downarrow$) &0.441&0.323& 0.412\\   
       
        \bottomrule
    \end{tabular}
\end{table}

\begin{table}[t]
    \caption{Impact of block partitioning.} 
\setlength{\tabcolsep}{1pt}
    \label{tab:partioning}
    \centering
    \scriptsize
    \begin{tabular}{lc|ccccccc}
        \toprule
        Baselines &Metrics$^\dagger$ & Size = 4 &Size= 7 &Size= 8 &Size = 14 & Size= 16 & Size = 28 & Size = 32\\
        \midrule
         \multirow{4}{*}{MINI-ImageNet}

        &SSIM ($\uparrow$) &  0.766  & 0.837 & 0.834  & 0.819 & 0.819 & 0.827 & 0.800\\
        &PSNR ($\uparrow$) &  26.679 & 28.335 & 28.321 & 27.737 & 27.885 & 28.228 & 27.484\\  
        &LPIPS ($\downarrow$) &  0.298  &0.218  & 0.212 & 0.227 & 0.224 & 0.220 & 0.254\\   
        \hline
        \multirow{4}{*}{CelebA-Subset}
        &SSIM ($\uparrow$) &  0.854 & 0.894 & 0.919 & 0.929 & 0.937 & 0.913 & 0.891\\
        &PSNR ($\uparrow$) &  30.236&31.810&33.332 & 34.063 & 34.664 & 32.981 & 31.809\\  
        &LPIPS ($\downarrow$) &0.234&0.208 &0.120 & 0.103 & 0.090 & 0.121 & 0.146\\   
        \hline
\multicolumn{2}{c|}{Parameters} & 2.22M & 2.32M & 2.37M & 2.78M & 2.96M & 4.58M & 5.32M \\
\multicolumn{2}{c|}{Memory time (s)} & 178.95 & 71.73 & 71.69 & 41.07 & 34.22 & 26.78 & 27.64 \\  
       
\bottomrule
    \end{tabular}
\end{table}
\begin{table*}[t]
    \caption{Impact of Secret Model Structure.} 
    \label{tab:selection}
    \centering
    \footnotesize
    \begin{tabular}{lc|cccccccc}
        \toprule
        Baselines &Metrics$^\dagger$&Random&Random with Constraints&Systematic&Layer-wise&Importance-based\\
        \midrule
        \multirow{4}{*}{\shortstack{CIFAR-10 }}
        &Leakage Rate ($\uparrow$) &100\% & 100\% & 100\%&100\%& 99.4\% \\
        &SSIM ($\uparrow$) &0.937& 0.867 & 0.881 & 0.895 & 0.727 &\\
        &PSNR ($\uparrow$) &28.251& 25.499 & 25.812 & 26.391 & 21.820 \\  
        &LPIPS ($\downarrow$) &0.083& 0.195 & 0.175 & 0.162 & 0.330 \\
        \cline{1-8}
        \multirow{4}{*}{\shortstack{CIFAR-100}}
        &Leakage Rate ($\uparrow$) &100\%& 99.6\% &100\% & 98.0\% & 100\%\\
        &SSIM ($\uparrow$) & 0.778 & 0.712 &0.824 & 0.666 & 0.779\\
        &PSNR ($\uparrow$) & 23.656 & 22.401 &24.376 & 21.567 & 23.637\\  
        &LPIPS ($\downarrow$) & 0.275 & 0.329 &0.232& 0.367 & 0.279\\   
        \cline{1-8}
        \multirow{4}{*}{MINI-ImageNet}
        &Leakage Rate ($\uparrow$) & 84.4\% & 82.0\% &92.2\% & 89.8\% & 91.4\% \\
        &SSIM ($\uparrow$) & 0.619 & 0.603 &0.659 & 0.636 & 0.646\\
        &PSNR ($\uparrow$) & 23.562 & 23.265 & 24.290& 23.874 & 24.047\\  
        &LPIPS ($\downarrow$) & 0.431 & 0.449 & 0.386 & 0.411 & 0.400\\   
        \cline{1-8}
        \multirow{4}{*}{CelebA-Subset}
        &Leakage Rate ($\uparrow$) & 99.2\% & 97.7\% & 96.9\% & 97.7\% &100\%\\
        &SSIM ($\uparrow$) & 0.621 & 0.625 & 0.617 & 0.620 &0.657\\
        &PSNR ($\uparrow$) & 23.883 & 23.781 & 23.496 & 23.801 &24.132\\  
        &LPIPS ($\downarrow$) &0.523& 0.511 & 0.510 & 0.519 &0.481\\   
       
        \bottomrule
    \end{tabular}
\end{table*}

\begin{table*}[t]
\caption{Comparison of \sys with Transpose \cite{amit2023transpose}, RtF\cite{fowl2021robbing} and LOKI\cite{zhao2023loki} on MINI-ImageNet and CELEBA-SUBSET datasets using FedAvg.
\\
\textit{The experiment was conducted five times, targeting different users in each trial, and the table shows the average performance results.}} 
    
    \label{tab:com_2}
    \centering
    \footnotesize
    \setlength{\tabcolsep}{3pt}
    \begin{tabular}{lc|cccccccc}
     \toprule

  \multicolumn{8}{c}{\color{black}MINI-ImageNet dataset}  \\

Baselines & Metrics$^\dagger$ & $N=8$ & $N=16$ & $N=24$ & $N=32$ & $N=48$  & $N=64$ \\
\midrule
\multirow{4}{*}{Transpose Attack} 
& Leakage ($\uparrow$) & 0.0 $\pm$ 0.0 & 0.2 $\pm$ 0.4 & 0.0 $\pm$ 0.0 & 0.0 $\pm$ 0.0 & 0.0 $\pm$ 0.0 & 0.0 $\pm$ 0.0  \\
& SSIM ($\uparrow$)    & 0.301 $\pm$ 0.028 & 0.247 $\pm$ 0.030 & 0.246 $\pm$ 0.014 & 0.247 $\pm$ 0.011  & 0.215 $\pm$ 0.018 & 0.193 $\pm$ 0.020   \\
& PSNR ($\uparrow$)    & 14.957 $\pm$ 1.698 & 13.700 $\pm$ 1.204 & 13.737 $\pm$ 0.992 & 13.842 $\pm$ 0.986 & 13.085 $\pm$ 0.783  & 12.508 $\pm$ 0.357  \\
& LPIPS ($\downarrow$) & 0.622 $\pm$ 0.015 & 0.657 $\pm$ 0.017 & 0.653 $\pm$ 0.004 & 0.653 $\pm$ 0.005 & 0.665 $\pm$ 0.010 & 0.688 $\pm$ 0.006  \\
\cline{1-8}
\multirow{4}{*}{RtF}
& Leakage ($\uparrow$) & 6.9 $\pm$ 0.2 & 13.0 $\pm$ 0.3 & 20.0 $\pm$ 0.1 
 & 26.2 $\pm$ 0.5  & 27.9 $\pm$ 1.0  & 52.1 $\pm$ 0.2  \\
& SSIM ($\uparrow$) &  0.844 $\pm$ 0.011  & 0.804 $\pm$ 0.015 & 0.816 $\pm$ 0.007 & 0.814 $\pm$ 0.013 & 0.850 $\pm$ 0.014 & 0.780 $\pm$ 0.004 \\
& PSNR ($\uparrow$) &  35.372 $\pm$ 0.870 & 34.094 $\pm$ 0.822 & 32.836 $\pm$ 0.374 & 33.407 $\pm$ 0.934 & 34.395 $\pm$ 0.850 & 30.052 $\pm$ 0.358 \\
& LPIPS ($\downarrow$) &  0.124 $\pm$ 0.005 & 0.161 $\pm$ 0.009 & 0.153 $\pm$ 0.005 & 0.156 $\pm$ 0.009 & 0.124 $\pm$ 0.008 & 0.182 $\pm$ 0.002 \\
\cline{1-8}
\multirow{4}{*}{LOKI}
& Leakage ($\uparrow$) & 6.0 $\pm$ 1.3 & 13.6 $\pm$ 1.2 & 19.8 $\pm$ 1.2 & 25.4 $\pm$ 3.5 & 40.8 $\pm$ 1.3 & 57.0 $\pm$ 3.0 \\
& SSIM ($\uparrow$) & 0.820 $\pm$ 0.173 & 0.882 $\pm$ 0.061 & 0.868 $\pm$ 0.052 & 0.816 $\pm$ 0.113 & 0.855 $\pm$ 0.028 & 0.866 $\pm$ 0.025 \\
& PSNR ($\uparrow$) & 53.756 $\pm$ 12.307 & 56.722 $\pm$ 6.192 & 57.763 $\pm$ 4.288 & 53.781 $\pm$ 10.556 & 54.719 $\pm$ 4.000 &57.071 $\pm$ 1.164 \\
& LPIPS ($\downarrow$) & 0.128 $\pm$ 0.127 & 0.083 $\pm$ 0.032 & 0.101 $\pm$ 0.045 & 0.141 $\pm$ 0.079 & 0.116 $\pm$ 0.025 & 0.102 $\pm$ 0.011\\

\cline{1-8}
\multirow{4}{*}{Ours} 
& Leakage ($\uparrow$) &\cellcolor[gray]{0.8}8.0 $\pm$ 0.0 &\cellcolor[gray]{0.8}16.0 $\pm$ 0.0 &\cellcolor[gray]{0.8}24.0 $\pm$ 0.0 &\cellcolor[gray]{0.8}32.0 $\pm$ 0.0 &\cellcolor[gray]{0.8}47.6 $\pm$ 0.8  &\cellcolor[gray]{0.8}61.4 $\pm$ 1.7 \\
& SSIM ($\uparrow$) &\cellcolor[gray]{0.8}0.962 $\pm$ 0.009 &\cellcolor[gray]{0.8}0.886 $\pm$ 0.007 &\cellcolor[gray]{0.8}0.837 $\pm$ 0.012 &\cellcolor[gray]{0.8}0.827 $\pm$ 0.015 &\cellcolor[gray]{0.8}0.755 $\pm$ 0.019 &\cellcolor[gray]{0.8}0.718 $\pm$ 0.015 \\
& PSNR ($\uparrow$) &\cellcolor[gray]{0.8}35.608 $\pm$ 0.846 &\cellcolor[gray]{0.8}30.185 $\pm$ 0.470 &\cellcolor[gray]{0.8}28.236 $\pm$ 0.658 &\cellcolor[gray]{0.8}28.228 $\pm$ 0.584 &\cellcolor[gray]{0.8}26.048 $\pm$ 0.298 &\cellcolor[gray]{0.8}25.128 $\pm$ 0.266 \\
& LPIPS ($\downarrow$) &\cellcolor[gray]{0.8}0.057 $\pm$ 0.013 &\cellcolor[gray]{0.8}0.153 $\pm$ 0.008 &\cellcolor[gray]{0.8}0.212 $\pm$ 0.012 
&\cellcolor[gray]{0.8}0.220 $\pm$ 0.012 &\cellcolor[gray]{0.8}0.296 $\pm$ 0.012 &\cellcolor[gray]{0.8}0.338 $\pm$ 0.009 \\

\bottomrule
 \multicolumn{8}{c}{\color{black}CELEBA-SUBSET dataset}  \\

Baselines & Metrics$^\dagger$ & $N=8$ & $N=16$ & $N=24$ & $N=32$ & $N=48$  & $N=64$ \\
\midrule
\multirow{4}{*}{Transpose Attack}

& Leakage ($\uparrow$) & 0.0 $\pm$ 0.0 & 1.4 $\pm$ 1.5 & 0.2 $\pm$ 0.4 & 0.4 $\pm$ 0.8 & 0.6 $\pm$ 0.4 & 0.6 $\pm$ 1.2 \\
& SSIM ($\uparrow$) & 0.410 $\pm$ 0.015 & 0.415 $\pm$ 0.030 & 0.388 $\pm$ 0.022 & 0.382 $\pm$ 0.021 & 0.367 $\pm$ 0.024 & 0.371 $\pm$ 0.027 \\
& PSNR ($\uparrow$) & 19.268 $\pm$ 0.759 & 18.967 $\pm$ 1.097 & 18.445 $\pm$ 0.835 & 17.918 $\pm$ 0.600 & 17.746 $\pm$ 0.572 & 17.720 $\pm$  0.772  \\
& LPIPS ($\downarrow$) & 0.573 $\pm$ 0.006 & 0.563 $\pm$ 0.013 & 0.575 $\pm$ 0.009 & 0.578 $\pm$ 0.011 & 0.583 $\pm$ 0.010 & 0.0586 $\pm$  0.015 \\
\cline{1-8}
\multirow{4}{*}{RtF}               

& Leakage ($\uparrow$) & 6.8 $\pm$ 0.2 & 14.3 $\pm$ 0.3 & 20.4 $\pm$ 0.8 & 25.8 $\pm$ 0.5 & 28.0 $\pm$ 0.9 & 52.9 $\pm$ 2.4 \\
& SSIM ($\uparrow$) &  0.833 $\pm$ 0.024 & 0.886 $\pm$ 0.010 & 0.816 $\pm$ 0.018 & 0.775 $\pm$ 0.009 & 0.841 $\pm$ 0.020 & 0.785 $\pm$ 0.024\\
& PSNR ($\uparrow$) & 33.992 $\pm$ 1.399 & 38.948 $\pm$ 0.185 & 32.074 $\pm$ 0.911 & 30.819 $\pm$ 0.426 & 32.542 $\pm$ 0.656 & 29.935 $\pm$ 0.775 \\
& LPIPS ($\downarrow$) &  0.145 $\pm$ 0.024 & 0.096 $\pm$ 0.005 & 0.168 $\pm$ 0.019 & 0.187 $\pm$ 0.006 & 0.146 $\pm$ 0.013 &  0.203 $\pm$ 0.019 \\
\cline{1-8}
\multirow{4}{*}{LOKI}
& Leakage ($\uparrow$) & 8.0 $\pm$ 0.0 & 12.0 $\pm$ 0.6  & 22.2 $\pm$ 0.7 & 28.2 $\pm$ 0.9 & 43.6 $\pm$ 0.8 &58.6 $\pm$ 1.6 \\

& SSIM ($\uparrow$) & 0.999 $\pm$ 0.000 & 0.755 $\pm$ 0.015 & 0.925 $\pm$ 0.021 & 0.896 $\pm$ 0.016 & 0.922 $\pm$ 0.009 & 0.905 $\pm$ 0.010\\
& PSNR ($\uparrow$) & 61.090 $\pm$ 0.000& 39.138 $\pm$ 0.425 & 60.148 $\pm$ 0.425 & 51.430 $\pm$ 0.470 & 55.590 $\pm$ 0.315 & 52.718 $\pm$ 0.643 \\
& LPIPS ($\downarrow$) & 0.001 $\pm$ 0.000 & 0.188 $\pm$ 0.012 & 0.061 $\pm$ 0.013 & 0.086 $\pm$ 0.007 & 0.061 $\pm$ 0.006 & 0.074 $\pm$ 0.007\\
\cline{1-8}
\multirow{4}{*}{Ours} 
& Leakage ($\uparrow$) &\cellcolor[gray]{0.8}8.0 $\pm$ 0.0 &\cellcolor[gray]{0.8}16.0 $\pm$ 0.0 &\cellcolor[gray]{0.8}24.0 $\pm$ 0.0 &  \cellcolor[gray]{0.8}32.0$\pm$ 0.0 &\cellcolor[gray]{0.8}48.0 $\pm$ 0.0 &\cellcolor[gray]{0.8}64.0 $\pm$ 0.0 \\
& SSIM ($\uparrow$)   &\cellcolor[gray]{0.8}0.974 $\pm$ 0.001 & \cellcolor[gray]{0.8} 0.944 $\pm$ 0.005 & \cellcolor[gray]{0.8} 0.918 $\pm$ 0.009 & \cellcolor[gray]{0.8} 0.913 $\pm$ 0.007 &  \cellcolor[gray]{0.8}0.886 $\pm$ 0.004 &  \cellcolor[gray]{0.8}0.865 $\pm$ 0.009 \\
& PSNR ($\uparrow$)    &  \cellcolor[gray]{0.8}39.546 $\pm$ 0.648 &  \cellcolor[gray]{0.8}35.618 $\pm$ 0.244 & \cellcolor[gray]{0.8} 33.551 $\pm$ 0.543 &  \cellcolor[gray]{0.8}32.981 $\pm$ 0.267 & \cellcolor[gray]{0.8} 31.683 $\pm$ 0.306 & \cellcolor[gray]{0.8} 30.727 $\pm$ 0.412 \\
& LPIPS ($\downarrow$) & \cellcolor[gray]{0.8} 0.039 $\pm$ 0.002 & \cellcolor[gray]{0.8} 0.080 $\pm$ 0.004 & \cellcolor[gray]{0.8} 0.115 $\pm$ 0.013 &  \cellcolor[gray]{0.8}0.121 $\pm$ 0.007 & \cellcolor[gray]{0.8}  0.157 $\pm$ 0.006 &  \cellcolor[gray]{0.8}0.185 $\pm$ 0.012 \\

\bottomrule

\end{tabular}
\begin{tablenotes}
    \centering
    \item  {\footnotesize $^\dagger$ ($\uparrow$) signifies that a higher value is preferable, while ($\downarrow$) indicates that a lower value is more desirable.}
\end{tablenotes}
\end{table*}

\begin{table*}[ht]
\caption{ {\color{black}Comparison of \sys with SEER \cite{garov2023hiding}, and Inverting \cite{geiping2020inverting} using FedSGD.}}
\vspace{0.1cm}
\label{tab:com_3}
\centering
\footnotesize
\begin{tabular}{lc|ccccccc}
\toprule

Dataset&Baselines &Metrics$^\dagger$&$N=16$&$N=32$&$N=64$&$N=128$&$N=256$&$N=512$\\
\midrule
\multirow{12}{*}{\shortstack{CIFAR-10}}&
\multirow{4}{*}{\shortstack{SEER}}
& Leakage ($\uparrow$) & 0.370 & 0.150 & 0.250 & 0.667 & 0.150 & 0 \\
&& SSIM ($\uparrow$)    & 0.492 & 0.468 & 0.481 & 0.501 & 0.486 & 0.306 \\
&& PSNR ($\uparrow$)    & 16.896 & 14.191 & 16.229 & 16.098 & 17.956 & 10.127 \\
&& LPIPS ($\downarrow$) & 0.399 & 0.476 & 0.419 & 0.401 & 0.386 & 0.669\\  
\cline{2-9} 
&\multirow{4}{*}{\shortstack{Inverting}}
&Leakage ($\uparrow$) & 0 & 0 & 0 & 0 & 0 & 0\\
&&SSIM ($\uparrow$) & 0.045 & 0.016 & 0.027 & 0.050 & 0.094 & 0.041 \\
&&PSNR ($\uparrow$) & 6.784 & 5.436 & 5.354 & 5.828 & 5.617 & 5.725\\  
&&LPIPS ($\downarrow$) & 0.673 & 0.686 & 0.690 & 0.702 & 0.714 & 0.704 \\
\cline{2-9}
&\multirow{4}{*}{\shortstack{Ours}}
& Leakage ($\uparrow$) &\cellcolor[gray]{0.8} 16 &\cellcolor[gray]{0.8} 32 & \cellcolor[gray]{0.8}64 & \cellcolor[gray]{0.8}128 & \cellcolor[gray]{0.8}256 & \cellcolor[gray]{0.8}511 \\
&& SSIM ($\uparrow$) & \cellcolor[gray]{0.8}0.997 & \cellcolor[gray]{0.8}0.998 &\cellcolor[gray]{0.8} 0.995 & \cellcolor[gray]{0.8}0.994 &\cellcolor[gray]{0.8} 0.938 & \cellcolor[gray]{0.8}0.729 \\
&& PSNR ($\uparrow$) & \cellcolor[gray]{0.8}37.530 & \cellcolor[gray]{0.8}37.754 & \cellcolor[gray]{0.8}37.183 & \cellcolor[gray]{0.8}36.773 & \cellcolor[gray]{0.8}29.967 & \cellcolor[gray]{0.8}22.657 \\
&& LPIPS ($\downarrow$) & \cellcolor[gray]{0.8}0.001 & \cellcolor[gray]{0.8}0.001 & \cellcolor[gray]{0.8}0.003 &\cellcolor[gray]{0.8} 0.004 &\cellcolor[gray]{0.8} 0.083 & \cellcolor[gray]{0.8}0.346 \\         \midrule
        
Dataset&Baselines & Metrics$^\dagger$ & $N=16$ & $N=32$ & $N=64$ & $N=128$ & $N=256$ & $N=512$ \\ 
\midrule
 \multirow{12}{*}{\shortstack{CIFAR-100}}&
\multirow{4}{*}{SEER} 
& Leakage ($\uparrow$) & 0.375 & 0.125 & 0.417 & 0.458 & 0.479 & 0.200 \\
&& SSIM ($\uparrow$)    & 0.481 & 0.439 & 0.482 & 0.495 & 0.496 & 0.462 \\
&& PSNR ($\uparrow$)    & 16.876 & 17.612 & 18.357 & 18.910 & 16.813 & 18.331 \\
&& LPIPS ($\downarrow$) & 0.438 & 0.434 & 0.408 & 0.385 & 0.391 & 0.402 \\
\cline{2-9}

&\multirow{4}{*}{Inverting} 
& Leakage ($\uparrow$) & 0 & 0 & 0 & 0 & 0 & 0 \\
&& SSIM ($\uparrow$)     & 0.065 & 0.061 & 0.050 & 0.062 & 0.049 & 0.082  \\
&& PSNR ($\uparrow$)     & 6.209 & 6.523 & 6.437 & 6.342 & 6.319 & 6.699 \\
&& LPIPS ($\downarrow$) & 0.636 & 0.614 & 0.658 & 0.663 & 0.676 & 0.673 \\
\cline{2-9}

&\multirow{4}{*}{Ours} 
& Leakage ($\uparrow$) & \cellcolor[gray]{0.8}16 & \cellcolor[gray]{0.8}32 & \cellcolor[gray]{0.8}64 & \cellcolor[gray]{0.8}128 & \cellcolor[gray]{0.8}256 & \cellcolor[gray]{0.8}511 \\
&& SSIM ($\uparrow$) & \cellcolor[gray]{0.8}0.996 & \cellcolor[gray]{0.8}0.998 &\cellcolor[gray]{0.8} 0.995 & \cellcolor[gray]{0.8}0.992 &\cellcolor[gray]{0.8}0.917 &\cellcolor[gray]{0.8} 0.718 \\
&& PSNR ($\uparrow$) & \cellcolor[gray]{0.8}37.157 & \cellcolor[gray]{0.8}41.566 &\cellcolor[gray]{0.8} 38.239 &\cellcolor[gray]{0.8} 35.794 & \cellcolor[gray]{0.8}29.107 &\cellcolor[gray]{0.8} 23.095 \\
&& LPIPS ($\downarrow$) & \cellcolor[gray]{0.8}0.001 &\cellcolor[gray]{0.8} 0.001 &\cellcolor[gray]{0.8} 0.002 & \cellcolor[gray]{0.8}0.006 & \cellcolor[gray]{0.8}0.108 & \cellcolor[gray]{0.8}0.331  \\
 \toprule
Dataset&Baselines &Metrics$^\dagger$&$N=8$&$N=16$&$N=24$&$N=32$&$N=48$&$N=64$\\

\midrule
\multirow{12}{*}{MINI-ImageNet}&
\multirow{4}{*}{SEER} 
& Leakage ($\uparrow$) & 0 & 0 & 0 & 0 & 0 & 0 \\
&& SSIM ($\uparrow$)    & 0.288 & 0.275 & 0.270 & 0.303 & 0.284 & 0.263 \\
&& PSNR ($\uparrow$)    & 23.242 & 21.256 & 19.263 & 18.137 & 15.944 & 14.342 \\
&& LPIPS ($\downarrow$) & 0.378 & 0.493 & 0.550 & 0.607 & 0.669 & 0.710 \\
\cline{2-9}
&\multirow{4}{*}{Inverting} 
& Leakage ($\uparrow$) & 0  & 0 & 0 & 0 & 0 & 0 \\
&& SSIM ($\uparrow$) & 0.029 & 0.035 & 0.068 & 0.048 & 0.033 & 0.036 \\
&& PSNR ($\uparrow$) & 6.063 & 6.588 & 6.754 & 6.710 & 6.712 & 6.729 \\
&& LPIPS ($\downarrow$) & 0.757 & 0.754 & 0.704 &  0.710 & 0.729 & 0.716 \\
\cline{2-9}
&\multirow{4}{*}{Ours} 
& Leakage ($\uparrow$) & \cellcolor[gray]{0.8}8 & \cellcolor[gray]{0.8}16 & \cellcolor[gray]{0.8}24 &\cellcolor[gray]{0.8} 32 &\cellcolor[gray]{0.8} 47 & \cellcolor[gray]{0.8}60 \\
&& SSIM ($\uparrow$) &\cellcolor[gray]{0.8} 0.884 &\cellcolor[gray]{0.8} 0.773 &\cellcolor[gray]{0.8} 0.767 &\cellcolor[gray]{0.8} 0.740 &\cellcolor[gray]{0.8} 0.676 &\cellcolor[gray]{0.8} 0.651 \\
&& PSNR ($\uparrow$) &  \cellcolor[gray]{0.8}29.837 & \cellcolor[gray]{0.8}26.882 &\cellcolor[gray]{0.8} 26.308 & \cellcolor[gray]{0.8}26.065 & \cellcolor[gray]{0.8}24.959 & \cellcolor[gray]{0.8}24.101 \\
&& LPIPS ($\downarrow$) & \cellcolor[gray]{0.8}0.212 & \cellcolor[gray]{0.8}0.298 & \cellcolor[gray]{0.8}0.312 & \cellcolor[gray]{0.8}0.339 &\cellcolor[gray]{0.8} 0.392 & \cellcolor[gray]{0.8}0.421 \\
\toprule
 Dataset&Baselines &Metrics$^\dagger$&$N=8$&$N=16$&$N=24$&$N=32$&$N=48$&$N=64$\\
        \midrule

\multirow{12}{*}{CelebA-Subset}&
\multirow{4}{*}{SEER} 
& Leakage ($\uparrow$) & 0 & 0 & 0 & 0 & 0 & 0 \\
&& SSIM ($\uparrow$)    & 0.359 & 0.361 & 0.345 & 0.347 & 0.351 & 0.327 \\
&& PSNR ($\uparrow$)    & 22.050 & 23.113 & 21.600 & 23.191 & 25.388 & 24.547 \\
&& LPIPS ($\downarrow$) & 0.277 & 0.266 & 0.288 & 0.303 & 0.351 & 0.390 \\
\cline{2-9}
&\multirow{4}{*}{Inverting} 
& Leakage ($\uparrow$) & 0 & 0 & 0 & 0 & 0 & 0 \\
&& SSIM ($\uparrow$) & 0.041 & 0.038 & 0.061 & 0.036 & 0.059 & 0.033 \\
&& PSNR ($\uparrow$) & 6.555 & 6.736 & 6.615 & 6.628 & 6.778 & 6.784 \\
&& LPIPS ($\downarrow$) & 0.773 & 0.774 & 0.759 & 0.783 & 0.778 & 0.784 \\
\cline{2-9}
&\multirow{4}{*}{Ours} 
& Leakage ($\uparrow$) & \cellcolor[gray]{0.8}8 & \cellcolor[gray]{0.8}16 & \cellcolor[gray]{0.8}24 &\cellcolor[gray]{0.8} 32 & \cellcolor[gray]{0.8}48 & \cellcolor[gray]{0.8}64 \\
&& SSIM ($\uparrow$) & \cellcolor[gray]{0.8}0.883 & \cellcolor[gray]{0.8}0.866 & \cellcolor[gray]{0.8}0.848 & \cellcolor[gray]{0.8}0.826 & \cellcolor[gray]{0.8}0.809 & \cellcolor[gray]{0.8}0.788 \\
&& PSNR ($\uparrow$) & \cellcolor[gray]{0.8}30.364 &\cellcolor[gray]{0.8}30.559 &\cellcolor[gray]{0.8}29.848 & \cellcolor[gray]{0.8}28.946 & \cellcolor[gray]{0.8}28.316 & \cellcolor[gray]{0.8}27.608 \\
&& LPIPS ($\downarrow$) &\cellcolor[gray]{0.8} 0.187 & \cellcolor[gray]{0.8}0.219 & \cellcolor[gray]{0.8}0.259 &\cellcolor[gray]{0.8} 0.286 &\cellcolor[gray]{0.8} 0.310 & \cellcolor[gray]{0.8}0.341 \\
\bottomrule

\end{tabular}
\end{table*}

\end{document}